\PassOptionsToPackage{table}{xcolor}
\documentclass[sigconf]{acmart}
\AtBeginDocument{%
  }

\setcopyright{acmlicensed}
\copyrightyear{2018}
\acmYear{2018}
\acmDOI{XXXXXXX.XXXXXXX}

\acmConference[MM '25]{Make sure to enter the correct
  conference title from your rights confirmation emai}{October 27-31,
  2025}{Dublin, Ireland.}
\acmISBN{978-1-4503-XXXX-X/2018/06}
\renewcommand\footnotetextcopyrightpermission[1]{}

\usepackage{amsmath}  
\everymath{\displaystyle}
\usepackage{bm}       
\usepackage{graphicx} 
\usepackage{xspace}   
\usepackage{multirow}
\usepackage{hyperref}
\usepackage{rotating}
\definecolor{tab_others}{RGB}{235, 235, 235}
\definecolor{tab_ours}{RGB}{225, 235, 246}
\usepackage{makecell}

\copyrightyear{2025}
\acmYear{2025}
\setcopyright{acmlicensed}\acmConference[MM '25]{Proceedings of the 33nd ACM International Conference on Multimedia}{October 27-31, 2025}{Dublin, Ireland}
\acmBooktitle{Proceedings of the 33nd ACM International Conference on Multimedia (MM '25), October 27-31, 2025, Dublin, Ireland}
\acmISBN{979-8-4007-0686-8/24/10}

\begin{document}

\title{Search is All You Need for Few-shot Anomaly Detection}
\author{Qishan Wang}
\authornote{These authors contributed equally to this work.}
\orcid{0000-0003-3463-9040}
\affiliation{%
  \department{Academy for Engineering and Technology}
  \institution{Fudan University}
  \country{Shanghai, China}
}
\email{qswang20@fudan.edu.cn}

\author{Jia Guo}
\authornotemark[1]
\authornote{Corresponding author}
\affiliation{%
  \department{School of Biomedical Engineering}
  \institution{Tsinghua University}
  \country{Beijing, China}}
\email{j-g24@mails.tsinghua.edu.cn}

\author{Shuyong Gao}
\affiliation{%
  \department{School of Computer Science}
  \institution{Fudan university}
  \country{Shanghai, China}
}
\email{sygao18@fudan.edu.cn}

\author{Haofen Wang}
\affiliation{%
 \department{College of Design \& Innovation}
 \institution{Tongji University}
\country{Shanghai, China}}
\email{carter.whfcarter@gmail.com}

\author{Li Xiong}
\affiliation{%
  \department{School of Physics and Mechanical \& Electrical Engineering}
  \institution{Hexi University}
  \country{Gansu, China}}
  \email{xl2025@hxu.edu.cn}

\author{Junjie Hu}
\affiliation{%
  \department{School of Computer Science}
  \institution{Fudan university}
  \country{Shanghai, China}
}
\email{23210240182@m.fudan.edu.cn}

\author{Hanqi Guo}
\affiliation{%
  \department{School of Computer Science}
  \institution{Fudan university}
  \country{Shanghai, China}
}
\email{hqguo23@m.fudan.edu.cn}

\author{Wenqiang Zhang}
\authornotemark[2]
\affiliation{%
  \department[0]{Engineering Research Center of AI\&Robotics, Ministry of Education}
  \department[1]{Academy for Engineering\&Technology}
  \institution{Fudan university}
  \country{Shanghai, China}
}
\email{wqzhang@fudan.edu.cn}


\begin{abstract}
Few-shot anomaly detection (FSAD) has emerged as a crucial yet challenging task in industrial inspection, where normal distribution modeling must be accomplished with only a few normal images. While existing approaches typically employ multi-modal foundation models combining language and vision modalities for prompt-guided anomaly detection, these methods often demand sophisticated prompt engineering and extensive manual tuning. In this paper, we demonstrate that a straightforward nearest-neighbor search framework can surpass state-of-the-art performance in both single-class and multi-class FSAD scenarios.
Our proposed method, VisionAD, consists of four simple yet essential components: (1) scalable vision foundation models that extract universal and discriminative features; (2) dual augmentation strategies - support augmentation to enhance feature matching adaptability and query augmentation to address the oversights of single-view prediction; (3) multi-layer feature integration that captures both low-frequency global context and high-frequency local details with minimal computational overhead;
and (4) a class-aware visual memory bank enabling efficient one-for-all multi-class detection.
Extensive evaluations across MVTec-AD, VisA, and Real-IAD benchmarks demonstrate VisionAD's exceptional performance.
Using only 1 normal images as support, our method achieves remarkable image-level AUROC scores of 97.4\%, 94.8\%, and 70.8\% respectively, outperforming current state-of-the-art approaches by significant margins (+1.6\%, +3.2\%, and +1.4\%).
The training-free nature and superior few-shot capabilities of VisionAD make it particularly appealing for real-world applications where samples are scarce or expensive to obtain.
Code is available at \href{https://github.com/Qiqigeww/VisionAD}{github.com/Qiqigeww/VisionAD}.
\end{abstract}

\ccsdesc[500]{Computing methodologies~Visual inspection}
\ccsdesc[500]{Computing methodologies~Anomaly detection}
\keywords{Anomaly detection, Few-shot Learning, Feature Representation, Vision Foundation}


\maketitle
\begin{sloppypar}

\section{Introduction}

Anomaly Detection (AD) is a critical task aimed at identifying anomalous samples in images that deviate from established normal patterns, with broad applications in fields such as industrial defect detection~\cite{10502267} and medical disease diagnosis~\cite{guo2023encoder}. Due to the unpredictability and scarcity of anomalies, along with inherent difficulties in collecting and annotating anomalous samples in real-world scenarios, traditional AD methods typically adopt an unsupervised paradigm, training only on accessible normal samples. However, obtaining sufficient normal samples can still be challenging, especially in scenarios such as the initial manufacturing stages of new products or situations where sample collection itself is difficult. As such, Few-Shot Anomaly Detection (FSAD) has been proposed to reduce annotation costs and novel anomaly detection issues, which holds significant promise for practical applications.

\begin{figure}[!t]
  \centering
  \includegraphics[width=\linewidth]{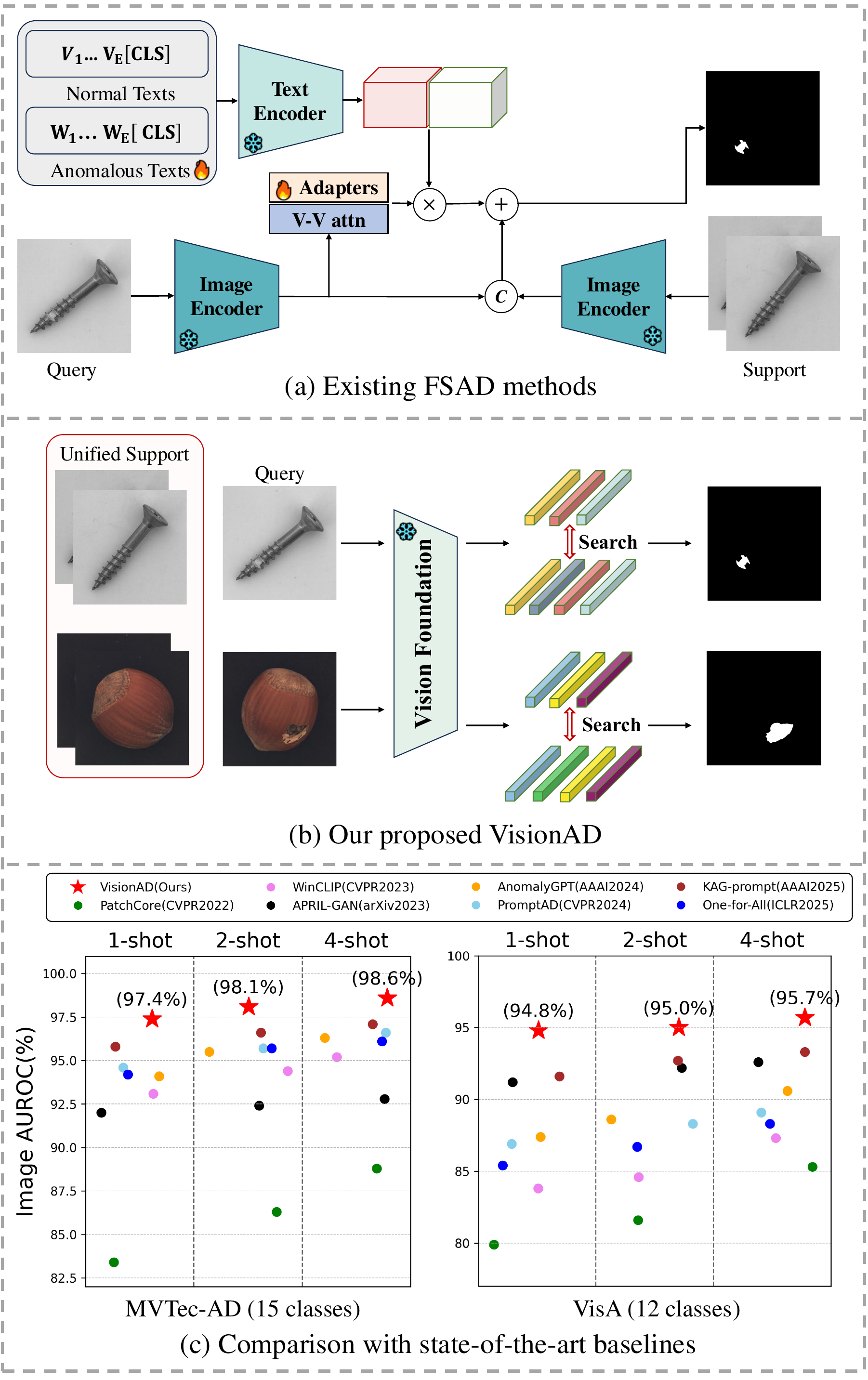}
  \caption{Comparisons between VisionAD and existing FSAD methods. (a) Existing FSAD models rely on complex manual or learnable text prompts and simple adapters or V-V attention for local features, resulting in a cumbersome "one-category-one-model" paradigm. (b) Our VisionAD, a plain, training-free vision-guided approach, generalizes effectively across multiple classes ("one-for-all" paradigm). (c) Comparison with previous SoTA methods on MVTec-AD~\cite{bergmann2019mvtec} and VisA~\cite{zou2022spot} across various settings, such as 1-shot, 2-shot, and 4-shot support images.}
  \label{motivation}
  \Description{}
\end{figure}

FSAD aims at localizing anomalous regions within a query image by leveraging a few normal support images belonging to the same object category. Recently, inspired by the remarkable zero-shot classification capabilities exhibited by vision-language foundation models, particularly CLIP~\cite{radford2021learning}, WinCLIP~\cite{jeong2023winclip} and subsequent studies~\cite{cao2024adaclip, zhou2023anomalyclip} have been proposed to exploit CLIP's powerful representation abilities to enhance the performance of low-shot anomaly detection models. Most existing methods fuse prompt-guided and vision-guided anomaly detection branches to obtain the final anomaly score. The vision-guided branch typically stores the nominal patch features into a visual memory bank, detecting anomalies in test samples based on high feature distances to their closest counterparts stored in memory. On the other hand, the prompt-guided anomaly detection branch generates anomaly maps by performing matrix operations between text and image features. For the text, numerous artificial prompts are usually manually designed~\cite{tao2024kernel} and/or learned~\cite{li2024promptad}, subsequently aggregated to yield normal or abnormal textual features. However, since the aforementioned process relies heavily on meticulous manual designs and/or partial learning strategies, it is burdensome, struggles to generalize to unseen anomalies, and requires bespoke model weights for each dataset category—known as the "one-category-one-model" paradigm (see Fig.~\ref{motivation}(a)). For the query image, as CLIP inherently struggles to capture fine-grained local features, existing methods typically resort to simple adapters~\cite{chen2023zero, gu2024anomalygpt} that map image features to the joint embedding space for comparison with textual features. Alternatively, Li et al.~\cite{li2023clip} employ a V-V attention mechanism to preserve local information within the CLIP token representations. Considering these intrinsic limitations of the CLIP-based model for FSAD, we are motivated to explore a plain yet effective, training-free approach that solely relies on vision-guided anomaly detection and can generalize across various classes, referred to as the "one-for-all" paradigm.

Inspired by the observation that few-shot anomaly detection is feasible for human annotators relying solely on visual features, we propose a visual-only approach, VisionAD, which performs nearest neighbor searches based on vision foundation models and a visual memory bank. VisionAD can handle various types of product anomalies without requiring domain-specific training data. Instead, it only needs a few normal samples from the target category during the testing phase to perform anomaly detection. Due to its simplicity and effectiveness, VisionAD can be easily deployed in industrial scenarios, as illustrated in Figure~\ref{motivation}(b).

Specifically, VisionAD introduces three straightforward yet crucial elements to address the critical issue of limited support data, which typically leads to insufficient patch representations, without increasing complexity or computational burden. First, intuitively, increasing the diversity of support data through simple data augmentation strategies can enhance the adaptability of feature matching. Augmentations, such as multiple rotations and flips applied to support images, can further align the shape and orientation of anomalies—crucial for performance but largely overlooked in previous studies. In addition to increasing the number of support samples, we propose to leverage the consistency and complementarity provided by pseudo multi-view data (i.e., different views of the same object are semantically consistent yet complementary) to mitigate potential oversights of details and interference from irrelevant local features caused by single-view predictions. Second, previous methods typically compare features from different layers separately and then fuse them, causing information isolation among layers. In contrast, we propose a multi-level feature fusion strategy that integrates intermediate-layer features to jointly capture long-range low-frequency and short-range high-frequency information before performing comparisons, thus effectively enhancing cross-scale feature interactions while minimizing bias towards ImageNet classes. Third, a class-aware visual memory bank is adopted within the one-for-all paradigm by leveraging the similarity between the class tokens of the test image and stored ones.

Extensive experimental results on the MVTec, VisA, and Real-IAD datasets show that the proposed VisionAD achieves unprecedented image-level AUROC scores of 97.4\%, 94.8\%, and 70.8\% under the 1-shot setting, respectively, surpassing previous state-of-the-art methods by a large margin, as illustrated in Figure~\ref{motivation}(c).

Our contributions are summarized as follows.
\begin{itemize}
    \item We propose VisionAD, a plain yet effective vision-only framework for FSAD that operates without complex prompt engineering or dataset-specific training, demonstrating that sophisticated language-vision fusion is not necessary for achieving superior performance in FSAD.
    \item We introduce three crucial components to address the limited support data challenge: dual augmentation strategy that combines support augmentation and pseudo multi-view transformation, multi-level feature fusion for capturing both global and local information, and class-aware visual memory banks enabling efficient one-for-all detection across categories.
    \item We conduct comprehensive experiments on popular benchmarks, demonstrating that VisionAD achieves state-of-the-art performance in few-shot anomaly detection, establishing its effectiveness and generalizability across various product categories and datasets.
\end{itemize}
\section{Related Works}
Few-shot anomaly detection is designed for scenarios where only a limited amount of normal data is available for training. In this case, normal samples may not fully capture the variability of normality. 
Many current studies focus on addressing this challenge and can be classified into two categories: CLIP-based (prompt-guided) FSAD and feature-matching-based (vision-guided) FSAD.

\textbf{CLIP-based FSAD.}
Since CLIP~\cite{radford2021learning} has demonstrated remarkable performance in zero-shot and few-shot classification, it has gained significant attention and is widely studied for its potential application in anomaly detection.
CLIP not only provides descriptive visual embeddings but also computes the similarity between text prompts and test images, as seen in works~\cite{jeong2023winclip,chen2023zero,gu2024anomalygpt,li2024promptad,tao2024kernel,lvone}.
WinCLIP~\cite{jeong2023winclip} designs manual text prompts for both normal and anomalous cases, utilizing sliding windows to extract and aggregate multiscale patch-based votes.
APRIL-GAN~\cite{chen2023zero} uses additional linear layers to align patch-level image features with textual features in order to generate anomaly maps.
AnomalyGPT\cite{gu2024anomalygpt} generates training data by simulating anomalous images and producing corresponding textual descriptions for each image.
PromptAD~\cite{li2024promptad} constructs a large number of negative samples through semantic concatenation. Additionally, it introduces the explicit abnormal edges module to control the margin between normal prompt features and anomaly prompt features.
KAG-prompt~\cite{tao2024kernel} proposes a kernel-aware hierarchical graph to capture cross-layer contextual information, leading to more accurate anomaly prediction.
One-for-All~\cite{lvone} learns a class-shared prompt generator to adaptively generate suitable prompts for each instance. In addition, it proposes a category-aware memory bank to retrieve valid similar features under the one-for-all paradigm.
However, these methods either require elaborate prompt engineering or fine-tuning of the prompt embeddings.
%

\textbf{Feature-matching-based FSAD.}
Patch feature matching research is based on the underlying idea of comparing the patch features of test samples with those of normal samples to compute anomaly scores~\cite{gu2024univad,luo2025exploring,roth2022towards,cohen2020sub,defard2021padim}.
UniVAD~\cite{gu2024univad} employs the Contextual Component Clustering (C3) module, Component-Aware Patch Matching (CAPM), and Graph-Enhanced Component Modeling (GECM) modules to detect anomalies at different semantic levels and across various domains.
INP-Former~\cite{luo2025exploring} extracts intrinsic normal prototypes directly from the test image and guides the Decoder to reconstruct only normal tokens, with reconstruction errors serving as anomaly scores.
PatchCore~\cite{roth2022towards} uses a maximally representative memory bank of nominal patch features to detect and segment anomalous data at test time.
SPADE~\cite{cohen2020sub} relies on the K nearest neighbors of pixel-level feature pyramids to detect and segment anomalies within images.
PaDiM~\cite{defard2021padim} uses a pretrained convolutional neural network for patch embedding and multivariate Gaussian distributions to obtain a probabilistic representation of the normal class.
However, these approaches are not designed for few-shot settings and lack simplicity and generality, so their performance may not meet the demands of manufacturing.

In this research, instead of using category tokens extracted from the Q-Former~\cite{lvone}, we leverage class token similarity to retrieve target patch features. Additionally, we propose a straightforward nearest-neighbor search framework, driven by advanced vision foundations, that can efficiently perform anomaly detection with a limited amount of data, thereby meeting the demands of manufacturing.
%

\begin{figure*}[t]
  \centering
  \includegraphics[width=\linewidth]{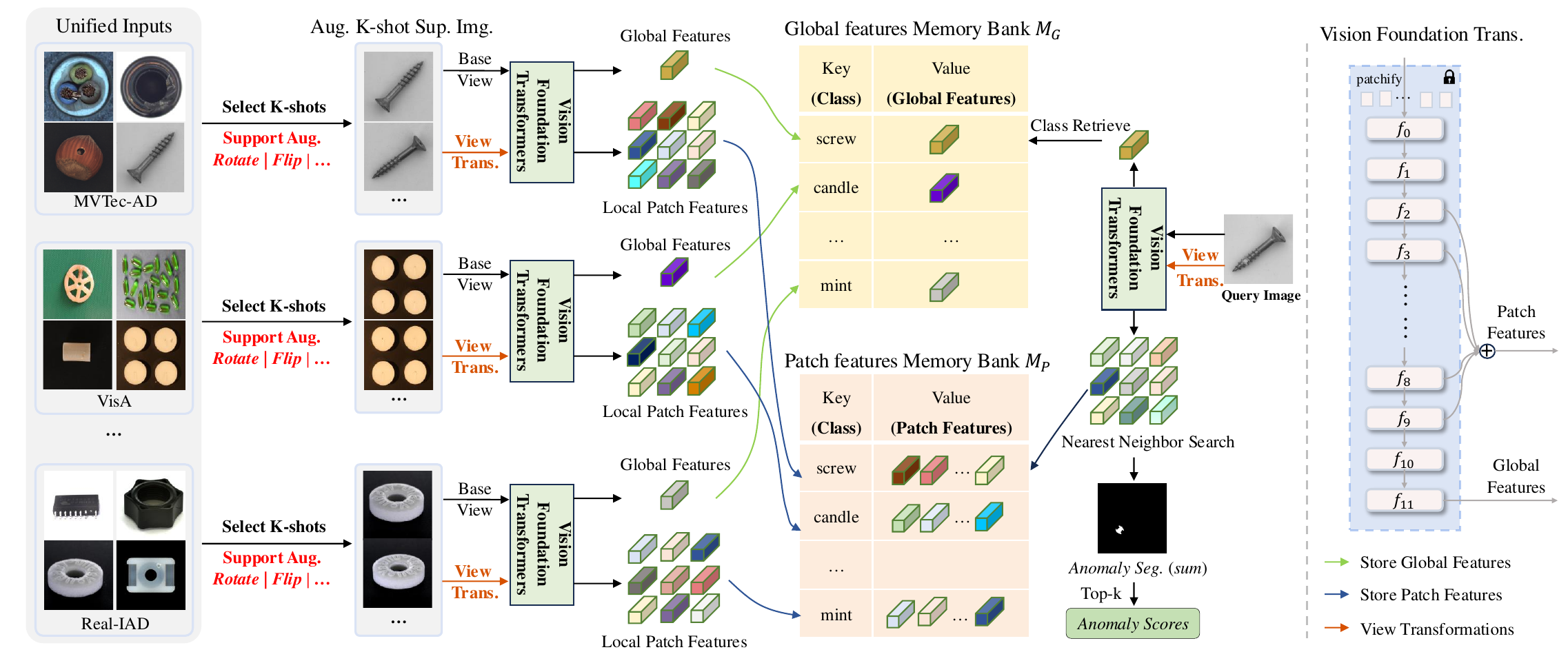}
  \caption{The overall framework of VisionAD. First, the selected k-shots of normal images in the support set are augmented, and an identical view transformation is applied to both the query and all support images (including augmented images). Next, patch features and global features for the reference normal images are extracted using a vision foundation model. These features are then stored in the global memory bank and patch memory bank, respectively. Meanwhile, the features and corresponding categories are combined into key–value pairs. During the testing stage, the category of the image is determined, and the scores from the base view and transformed views are fused to generate the anomaly detection and localization results in the designated patch memory bank.}
  \label{framework}
  \Description{}
\end{figure*}

\section{Method}

Given an FSAD dataset containing $N$ classes $\boldsymbol{C} = \{ \mathbf{C}_1, \mathbf{C}_2, \cdots, \mathbf{C}_N\}$, the general few-shot anomaly detection setting typically selects only one class $\mathbf{C}_{i}$ (where $i=1, 2, \cdots, N$ indicates the class number), using the one-category-one-model set $\chi_{i} = \{(x_{i, normal}^{Reference}), (x_{i, normal}^{Test}, x_{i, anomaly}^{Test})\}$. In contrast, a more practical one-for-all setting encompasses all classes $\boldsymbol{C}$, covering the all-class sets $\chi =  {\textstyle \sum_{i=1}^{N}\chi_{i}}$ within one unified model. It should be noted that under both settings, each class contains at most $K$ examples, with $K$ typically taking values of 1, 2, or 4. In this paper, we focus on leveraging the reference dataset $\boldsymbol{C}$ to detect and localize anomalies across various categories under the one-for-all paradigm.

\subsection{VisionAD Framework}
Since language-guided approaches require elaborate designs for training and often struggle to capture fine-grained local features, certain defects can only be detected via visual reference. For instance, "Metal Nut" in MVTecAD has an anomaly type labeled as "flipped upside-down," which can only be identified relatively from a normal image. Therefore, we propose VisionAD, a purely vision-based patch-matching framework that leverages vision foundation models and effective feature fusion to construct a memory bank M. This memory bank stores informative patch-level feature maps of nominal samples at multiple semantic scales. Then, we apply data augmentation to expand the memory bank and better align it with test samples. Identical augmentations are simultaneously applied to both query and reference images to obtain more informative multi-view anomaly scores. Additionally, we introduce a class-aware memory bank to retrieve category-specific support images, facilitating more precise anomaly detection under one-for-all setting. The overall framework of our method is illustrated in Figure~\ref{framework}.

\textbf{Vision Foundation Transformers.}
In order to extract rich features for both normal and anomalous samples, we adopt advanced visual foundation models—particularly ViT architectures pretrained on large-scale datasets~\cite{liu2021swin}. These models typically leverage contrastive learning (e.g., MoCov3~\cite{chen2021empirical}, DINO~\cite{caron2021emerging}), masked image modeling (e.g., MAE~\cite{he2022masked}, SimMIM~\cite{xie2022simmim}, BEiT~\cite{peng2022beit}), or their combinations(e.g., iBOT~\cite{zhou2021ibot}, DINOv2~\cite{oquab2023dinov2}), efficiently capturing both semantic and fine-grained information to generate universal and discriminative features. Specifically, we apply a general-purpose feature extractor $F$ to the input image $x_i$, generating a feature map of dimensions \( H_f \times W_f \times D_f \). To fully harness the benefits of pre-trained models and enable precise differentiation between normal and anomalous samples, we freeze the parameters of $F$ throughout the anomaly detection process, thereby preventing gradient back-propagation.

Recent studies have shown that pretrained models, whether supervised or self-supervised, tend to generate robust and universal features for anomaly detection tasks. In this paper, we examine the scaling behaviors of VisionAD by systematically analyzing Vision foundational Transformers. Our comprehensive evaluation investigates various aspects, including pre-training strategies (Figure~\ref{Change_backbone}) and model sizes (Table~\ref{tab:arch_scale}), as detailed in Section~\ref{ablation}. Striking a balance between detection performance and computational efficiency, we adopt the ViT-Large/14 model pretrained using DINOv2-Register~\cite{darcet2023vision} as the default extractor.

\subsection{Feature Fusion}
Inspired by WinCLIP, most vision-language FSAD models employ a vision-guided search branch that stores multi-layer patch embeddings in memory, compares corresponding layers between memory and query images, and aggregates the resulting anomaly maps into a final prediction~\cite{jeong2023winclip, zhou2023anomalyclip, chen2023zero} (Figure~\ref{feature_fusion}(a)). Intuitively, comparing more feature pairs allows the FSAD model to exploit richer information across layers for anomaly discrimination. However, due to misalignment between the query and support images, the similarity scores among low-level normal features tend to be underestimated, while high-level features—lacking localized nominal information and overly biased toward the pre-training dataset—yield skewed similarity scores~\cite{roth2022towards}. These discrepancies obscure the true discrepancy between local anomalous features and the memory representations. Utilizing local neighbourhood aggregation (e.g., PatchCore~\cite{roth2022towards} stores multiple intermediate patch-level features) enhances robustness to small spatial deviations without sacrificing spatial resolution (Figure~\ref{feature_fusion}(b)). However, feature maps from ResNet-like~\cite{he2016deep} architectures have limited receptive fields and struggle to capture long-range correlations among distant positions. Given the top-to-bottom consistency of columnar Transformer layers, we propose aggregating all feature maps of interest into a single group rather than enforcing strict layer-to-layer comparisons (Figure~\ref{feature_fusion}(c)). This strategy leverages richer, more diversified features for search and comparison, enabling the detection of subtle local anomalies. Furthermore, since shallow layer features contain low-level visual characteristics critical for precise localization, we further partition the features into low-semantic-level and high-semantic-level groups (Figure~\ref{feature_fusion}(d)).

\subsection{Support Aug. and Pseudo Multi-View}

\begin{figure}[t]
  \centering
  \includegraphics[width=\linewidth]{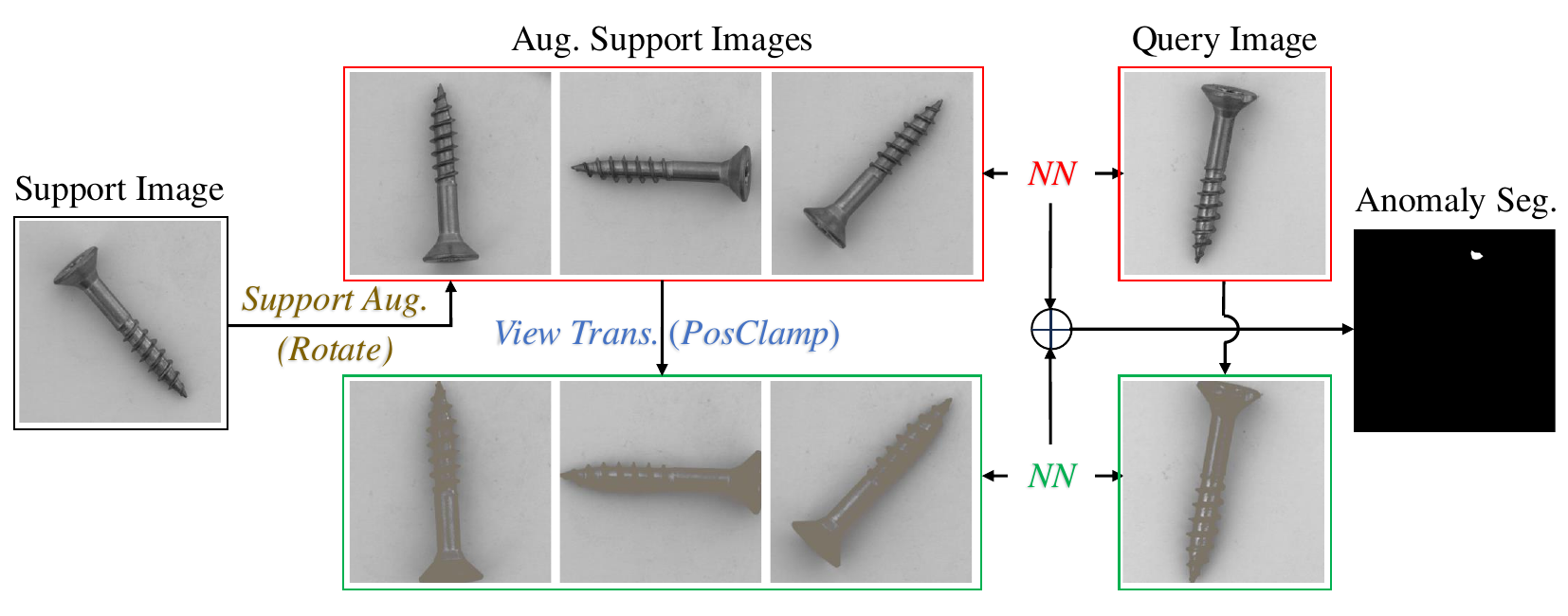}
  \caption{Support Enhancement and Pseudo Multi-View for Anomaly Detection.}
  \label{screw_metal}
\end{figure}

\textbf{Support Augmentation.} A major challenge faced by FASD is the scarcity of normal samples in the support set, often with fewer than 8 samples available. This results in the difficulty of storing normal features in memory that adequately cover the query features. Previous studies, such as UniVAD~\cite{gu2024univad} and One-for-All~\cite{lvone}, have focused on improving the accuracy of feature matching; however, the low matching quality is often caused by misalignment between the query and support images. A direct solution to this issue is to increase the diversity of the support data. We observe that most industrial images under certain categories are highly similar and can be transformed into one another through simple data augmentation, such as rotation, translation, and flipping, as demonstrated by the screw in the Figure~\ref{screw_metal}. Consequently, our natural inclination is to acquire additional data through augmentation strategies like rotation and flipping. This allows the feature memory bank to store more useful features.
%

\textbf{Pseudo Multi-View.} We find that single-view predictions may overlook important details and focus on trivial features. Drawing inspiration from the complementarity and consistency inherent in multi-view data, we propose a novel approach that simultaneously transforms both query and support images through identical projections. This technique enables the extraction of both shared semantic information and view-specific features, leading to more robust and accurate predictions.
The "view" represents not only the camera-shooting perspective, but also the distinct representations of the same sample in multi-view learning~\cite{yan2021deep}. Specifically, we find that simple projection methods, such as image thresholding and flipping (followed by anomaly map inversion), yield substantial performance improvements, as illustrated in Figure~\ref{screw_metal}. These transformations can simulate different viewing conditions: thresholding mimics camera parameter adjustments, while flipping emulates alternative viewing angles.

\subsection{Category-Indexed Memory Bank}
Under the one-for-all paradigm, mixing patch features from multiple categories within a single memory bank can complicate the nearest neighbor search and increase computation time. Additionally, regional similarities between images from different categories may lead to undesired feature matching. To address these issues, patch features \( \mathbf{P}\left ( \mathbf{F} \left ( x_i \right ) \right ) \) extracted from normal image \( x_i \) in the support set are stored in a dedicated patch feature memory bank \( \mathbf{M}_p \), where they are paired with the corresponding category label \( \mathbf{C}_i \). Similarly, global features \( \mathbf{G}_i \) extracted from normal image \( x_i \) in the support set are stored in a separate global feature memory bank \( \mathbf{M}_G \), with key-value pairs formed based on their category \( \mathbf{C}_i \).

The category-indexed patch and global feature memory banks \( \mathbf{M}_P \) and \( \mathbf{M}_G \) are defined as:
\begin{equation}\label{mp}
    \mathbf{M}_P = \left\{ \left[ \mathbf{C}_1; \mathbf{P}\left ( \mathbf{F} \left ( x_1 \right ) \right ) \right],\ldots, \left[ \mathbf{C}_N; \mathbf{P}\left ( \mathbf{F} \left ( x_N \right ) \right ) \right] \right\}
\end{equation}
\begin{equation}\label{mg}
    \mathbf{M}_G = \left\{ \left[ \mathbf{C}_1; \mathbf{G}\left ( \mathbf{F} \left ( x_1 \right ) \right ) \right],\ldots, \left[ \mathbf{C}_N; \mathbf{G}\left ( \mathbf{F} \left ( x_N \right ) \right ) \right] \right\}
\end{equation}

During the testing phase, the category of the test image is first retrieved by measuring the similarity between the class token \( \mathbf{G}_s \) of the test image and the stored tokens. The corresponding patch features can then be retrieved through the category, enabling anomaly detection and localization, as shown in Figure~\ref{framework}. 

\subsection{Anomaly Detection and Segmentation}
First, we randomly select one sample from the k-shot normal samples of each class and construct a global feature memory bank, storing them as key-value pairs based on their respective categories (Equation~\eqref{mg}). Then, we expand the support set using augmentation strategies such as rotation and flipping. Their patch features are extracted using a vision foundation model and stored in the patch feature memory bank (Equation~\eqref{mp}).

In the testing phase, we first determine the category of the test image by measuring the distance between its class token and those stored in $\mathbf{M}_G$. The category corresponding to the global feature with the smallest distance is then selected as the category of the test image:
\begin{equation}
    \mathbf{C} = \min \mathcal{D}\left( \mathbf{G}\left( \mathbf{F}\left(x^{\text{test}}\right) \right), \mathbf{M}_G \right)
\end{equation}

Then, the localization result $Score_{pix\_ori}$ is obtained by calculating the distance between the query patch and the most similar corresponding patch in \( \mathbf{M}_p \):
\begin{equation}
    Score_{pix\_ori} = \operatorname{Up}\left(1 - \max_{m_j \in M_P} \left( \mathbf{P}_i\left( \mathbf{F}\left(x^{\text{test}}\right) \right) \cdot m_j \right)\right)
\end{equation}

Similarly, identical view transformation is applied to both the query and all reference images (including augmented ones) to obtain the localization result $Score_{pix\_view}$.

The final result of anomaly localization is:
\begin{equation}
    Score_{pix} = Score_{pix\_ori} + Score_{pix\_view}
\end{equation}

For image-level anomaly score, to retain potentially important information beyond the global maximum, we use the average of the top-$k$ highest values (top $1\%$ pixels in an anomaly map) in the prediction map as our metric. This strategy effectively aggregates multiple significant anomaly signals present in the prediction map. The image-level anomaly score is:
\begin{equation}
    Score_{img} = \operatorname{Mean}(\operatorname{Top}\text{-}k(Score_{pix}))
\end{equation}

\begin{figure}[t]
  \centering
  \includegraphics[width=\linewidth]{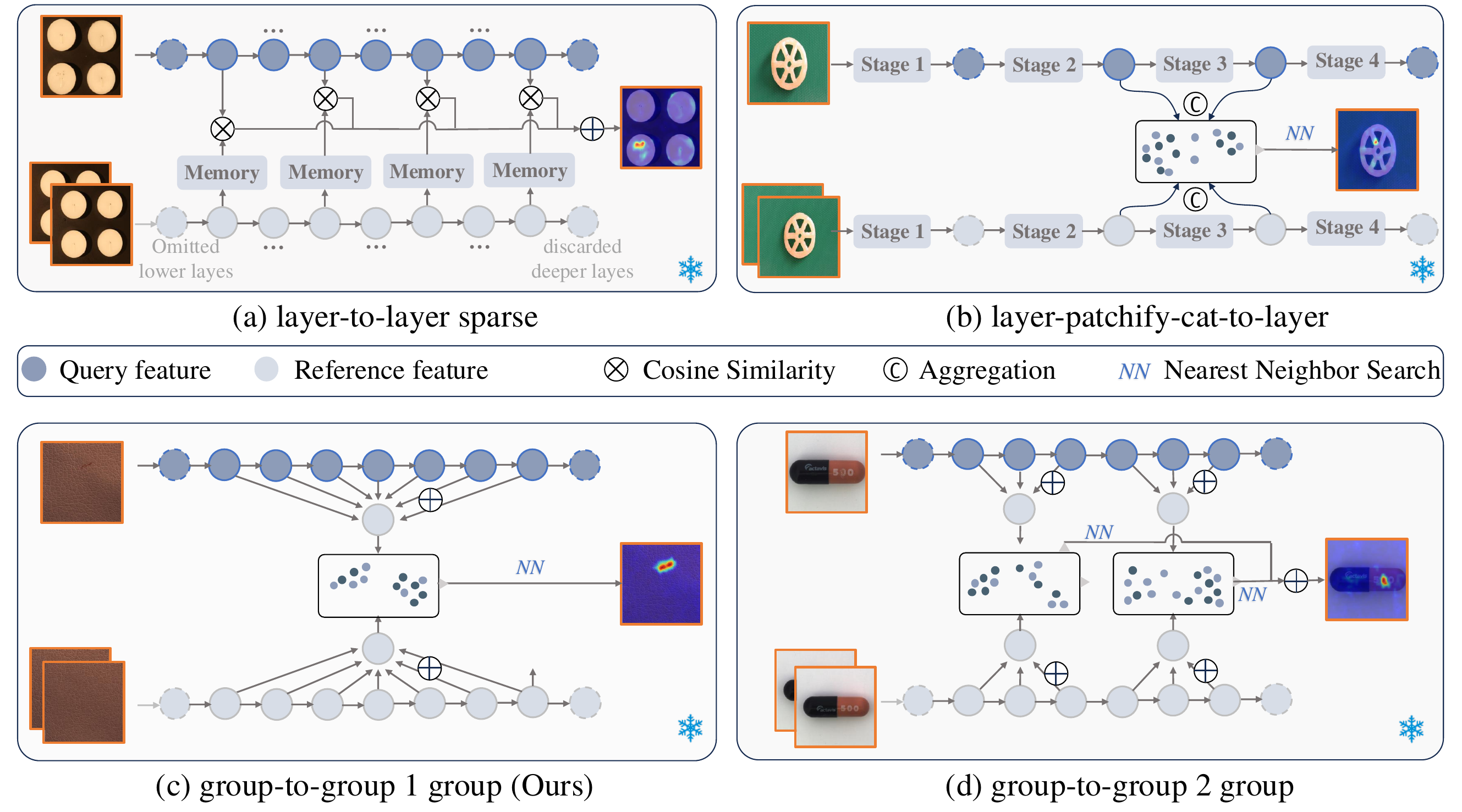}
  \caption{Schemes of Feature Fusion. (a) Layer-to-layer (sparse). (b) layer-patchify-cat-to-layer. (c) group-to-group, 1-group (Ours). (d) group-to-group, 2-group.}
  \label{feature_fusion}
  \Description{}
\end{figure}
\section{Experiments}
\subsection{Experimental Settings}

\textbf{Datasets.}
\textbf{MVTec-AD} \cite{bergmann2019mvtec} consists of 15 object categories, including 5 texture classes and 10 object classes, with 3,629 normal images in the training set and 1,725 images in the test set (467 normal and 1,258 anomalous).
\textbf{VisA} \cite{zou2022spot} contains 12 object categories. With the official data split, the training set includes 8,659 normal images, and the test set includes 2,162 images (962 normal and 1,200 anomalous).
\textbf{Real-IAD} \cite{wang2024real} is a large-scale, recently released dataset for unsupervised anomaly detection, covering 30 distinct object categories. The official split, which includes all views, results in 36,465 normal images in the training set and 114,585 images in the test set (63,256 normal and 51,329 anomalous).

\textbf{Metrics.}
Consistent with prior studies~\cite{guo2024dinomaly, zhang2023exploring}, we evaluate anomaly detection and localization using metrics such as the Area Under the Receiver Operating Characteristic Curve (AUROC). Additionally, we employ the Area Under the Precision-Recall Curve (AUPR) for anomaly detection and the Per-Region Overlap (PRO) for anomaly localization to provide a more comprehensive analysis of the model's performance.

\textbf{Implementation Details.}
In the few-normal-shot setting, no further training is performed on the anomaly detection datasets for VisionAD. By default, we use the ViT-Base/14 (patch size = 14) pre-trained by DINOv2-R \cite{darcet2023vision} as our image encoder, with its parameters frozen. The input image is first resized to $448^2$ and then center-cropped to $392^2$, ensuring the feature map ($28^2$) is sufficiently large for anomaly localization. As previously mentioned, we utilize the middle 8 layers of the 12-layer ViT-Base for feature fusion. ViT-Small, which also has 12 layers, follows the same selection. ViT-Large, with 24 layers, uses layers [4,6,8,...,19] for fusion (with indexing starting from 0).

\begin{table*}[t]
  \centering
    \caption{Performance comparisons of the FSAD methods on the MVTecAD and VisA datasets across different few-shot settings. We report the mean and standard deviation over 5 random seeds for each measurement. Bold indicates the best performance and underlining indicates sub-optimal results.}
    \setlength{\tabcolsep}{1.0mm}
    \resizebox{0.95\linewidth}{!}{
    \begin{tabular}{cccccccccccccc}
    \toprule
    \multirow{2}{*}{Setup} & \multirow{2}{*}{Method} & \multirow{2}{*}{Public} & \multicolumn{4}{c}{MVTecAD} & \multicolumn{4}{c}{VisA}& \multirow{2}{*}{avg} \\
\cmidrule(r){4-7} \cmidrule(r){8-11}  &       &       & AUROC & AUPR & pAUROC & PRO & AUROC & AUPR & pAUROC & PRO &\\
    \midrule
    \multirow{8}{*}{1-shot} & PatchCore~\cite{roth2022towards} &  CVPR2022 & 83.4  & 92.2   & 92.0  & 79.7  & 79.9 & 82.8  & 95.4  & 80.5    &85.7 \\
          & WinCLIP~\cite{jeong2023winclip} & CVPR2023 & 93.1  & 96.5   & 95.2  & 87.1   & 83.8 & 85.1 & 96.4  & 85.1   & 90.3  \\
          & APRIL-GAN~\cite{chen2023zero} & arXiv2023 & 92.0    & 95.8    & 95.1    &90.6    & 91.2  &  \underline{93.3}    & 96.0  & \underline{90.0}  & 93.0 \\
          & AnomalyGPT~\cite{gu2024anomalygpt} & AAAI2024 & 94.1  & 95.9    & 95.3 & 89.5   & 87.4  & 88.7  & 96.2  & 82.9  & 91.3 \\
          & PromptAD~\cite{li2024promptad} & CVPR2024 & 94.6    & 97.1  & 95.9  & 87.9  & 86.9  & 88.4  & 96.7  & 85.1  & 91.6 \\
          & KAG-prompt~\cite{tao2024kernel} & AAAI2025 & \underline{95.8}   & \underline{98.1} & 96.2   & \underline{90.8}  & \underline{91.6}  & 93.2 & \underline{97.0}  & 85.2   & \underline{93.5} \\
          & One-for-All~\cite{lvone} & ICLR2025 & 94.2 & 97.2   & \textbf{96.4}  & 89.8  & 85.4  & 87.5  & 96.9  & 87.3  & 91.8 \\          
\cmidrule{2-12}          & \textbf{VisionAD (ours)} & -     & \textbf{97.4\footnotesize{$\pm$0.4}} & \textbf{99.0\footnotesize{$\pm$0.2}}  & \underline{96.2\footnotesize{$\pm$0.2}} & \textbf{92.5\footnotesize{$\pm$0.3}}  & \textbf{94.8\footnotesize{$\pm$1.0}} & \textbf{95.0\footnotesize{$\pm$0.9}}  & \textbf{97.6\footnotesize{$\pm$0.3}}  & \textbf{91.6\footnotesize{$\pm$0.8}}  & \textbf{95.5} \\
    \midrule
    \multirow{8}{*}{2-shot} & PatchCore~\cite{roth2022towards} & CVPR2022 & 86.3 & 93.8  & 93.3  & 82.3  & 81.6  & 84.8  & 96.1  & 82.6  & 87.6 \\
          & WinCLIP~\cite{jeong2023winclip} & CVPR2023 & 94.4   & 97.0  & 96.0  & 88.4  & 84.6  & 85.8  & 96.8   & 86.2 & 91.1 \\
          & APRIL-GAN~\cite{chen2023zero} & arXiv2023 & 92.4    & 96.0  & 95.5  & \underline{91.3}  & 92.2  & \underline{94.2}  & 96.2  & \underline{90.1}  & 93.5 \\
          & AnomalyGPT ~\cite{gu2024anomalygpt} & AAAI2024 & 95.5   & 96.8  & 95.6  & 90.0  & 88.6  & 89.0  & 96.4  & 83.4  & 91.9 \\
          & PromptAD~\cite{tao2024kernel} & CVPR2024 & 95.7 & 97.9  & 96.2  & 88.5 & 88.3   & 90.0  & 97.1  & 85.8  & 92.4 \\
          & KAG-prompt~\cite{tao2024kernel} & AAAI2025 & \underline{96.6}   & \underline{98.5}  & 96.5  & 91.1   & \underline{92.7} & \underline{94.2}   & \underline{97.4}  & 86.7   & \underline{94.2} \\
          & One-for-All~\cite{lvone} & ICLR2025 & 95.7  & 97.9  & \textbf{96.7}  & 90.3 & 86.7   & 88.6   & 97.2  & 87.9 & 92.6 \\             
\cmidrule{2-12}          & \textbf{VisionAD (ours)} & -     & \textbf{98.1\footnotesize{$\pm$0.3}} & \textbf{99.3\footnotesize{$\pm$0.1}}  & \underline{96.6\footnotesize{$\pm$0.1}} & \textbf{93.2\footnotesize{$\pm$0.2}} & \textbf{95.0\footnotesize{$\pm$0.3}}    & \textbf{95.2\footnotesize{$\pm$0.6}} & \textbf{97.7\footnotesize{$\pm$0.0}} & \textbf{91.8\footnotesize{$\pm$0.3}} & \textbf{95.9} \\
    \midrule
    \multirow{8}{*}{4-shot} & PatchCore~\cite{roth2022towards} & CVPR2022 & 88.8 &94.5   &94.3   & 84.3  & 85.3  & 87.5  & 96.8  & 84.9  & 89.6 \\
          & WinCLIP~\cite{jeong2023winclip} & CVPR2023 & 95.2  & 97.3  & 96.2  & 89.0   & 87.3  & 88.8  & 97.2  & 87.6  & 92.3 \\
          & APRIL-GAN~\cite{chen2023zero} & arXiv2023 & 92.8  & 96.3   & 95.9  & \underline{91.8} & 92.6    & 94.5  & 96.2  & \underline{90.2}  & 93.8 \\
          & AnomalyGPT~\cite{gu2024anomalygpt} & AAAI2024 & 96.3    & 97.6  & 96.2  & 90.7  & 90.6  & 91.3  & 96.7  & 84.6  & 93.0 \\
          & PromptAD~\cite{tao2024kernel} & CVPR2024 & 96.6  & 98.5 & 96.5  & 90.5  & 89.1  & 90.8  & 97.4  & 86.2  & 93.2 \\
          & KAG-prompt~\cite{tao2024kernel} & AAAI2025 & \underline{97.1}   & \underline{98.8}  & 96.7   & 91.4  & \underline{93.3}  & \underline{94.6} & \textbf{97.7}  & 87.6   & \underline{94.7} \\
          & One-for-All~\cite{lvone} & ICLR2025 & 96.1 & 98.1   & \textbf{97.0}  & 91.2  & 88.3  & 89.6  & 97.4  & 88.3  & 93.2 \\             
\cmidrule{2-12}          & \textbf{VisionAD (ours)} & -     & \textbf{98.6\footnotesize{$\pm$0.1}} & \textbf{99.5\footnotesize{$\pm$0.1}}  & \underline{96.9\footnotesize{$\pm$0.1}} & \textbf{93.7\footnotesize{$\pm$0.1}} & \textbf{95.7\footnotesize{$\pm$0.3}}    & \textbf{95.9\footnotesize{$\pm$0.3}} & \textbf{98.0\footnotesize{$\pm$0.0}} & \textbf{92.5\footnotesize{$\pm$0.2}} & \textbf{96.4} \\
    \bottomrule
    \end{tabular}%
    }
  \label{mvtec-visa}
\end{table*}

\begin{table}[t]
  \centering
    \caption{Performance comparisons of the FSAD methods on the Real-IAD datasets across different few-shot settings.
    }
    \setlength{\tabcolsep}{1mm}
    \resizebox{\linewidth}{!}{
    \begin{tabular}{cccccccccc}
    \toprule
    \multirow{2}{*}{Setup} & \multirow{2}{*}{Method} & \multirow{2}{*}{Public} & \multicolumn{4}{c}{Real-IAD} & \multirow{2}{*}{avg} \\
\cmidrule(r){4-7}  &       &       & AUROC & AUPR & pAUROC & PRO &\\
    \midrule
    \multirow{7}{*}{1-shot} & SPADE~\cite{cohen2020sub} &  arXiv2020 & 51.2  & 45.6   & 59.5  & 19.3    & 43.9 \\
    & PaDiM~\cite{defard2021padim} &  ICPR2020 & 52.9  & 47.4   & 84.9  & 52.7    & 59.5 \\
    & PatchCore~\cite{roth2022towards} & CVPR2022 & 59.3 & 55.8  & 89.6  & 60.5  & 66.3 \\
    & WinCLIP~\cite{jeong2023winclip} & CVPR2023 & \underline{69.4}  & 56.8   & 91.9  & 71.0   & 72.3  \\
    & PromptAD~\cite{li2024promptad} & CVPR2024 & 52.2   & 41.6  & 84.9  & 58.4  & 59.3 \\
    & INP-Former~\cite{luo2025exploring} & CVPR2025 & 67.5   & \underline{63.1}  & \underline{94.9}  & \underline{81.8}  & \underline{76.8} \\
\cmidrule{2-8}          & \textbf{VisionAD (ours)} & -     & \textbf{70.8\footnotesize{$\pm$1.2}} & \textbf{66.7\footnotesize{$\pm$1.5}}  & \textbf{95.7\footnotesize{$\pm$0.3}} & \textbf{84.9\footnotesize{$\pm$0.7}}  & \textbf{79.5} \\
    \midrule    
    \multirow{7}{*}{2-shot} & SPADE~\cite{cohen2020sub} &  arXiv2020 & 50.9  & 45.5   & 59.5  & 19.2    & 43.8 \\
    & PaDiM~\cite{defard2021padim} &  ICPR2020 & 55.9  & 49.6   & 88.5  & 61.6    & 63.9 \\
    & PatchCore~\cite{roth2022towards} & CVPR2022 & 63.3 & 59.7  & 92.0  & 66.1  & 70.3 \\
    & WinCLIP~\cite{jeong2023winclip} & CVPR2023 & \underline{70.9}  & 58.7   & 93.2  & 74.7   & 74.4  \\
    & PromptAD~\cite{li2024promptad} & CVPR2024 & 57.7   & 41.1  & 86.4  & 61.0  & 61.6 \\ 
    & INP-Former~\cite{luo2025exploring} & CVPR2025 & 70.6   & \underline{66.1}  & \underline{96.0}  & \underline{83.8}  & \underline{79.1} \\
\cmidrule{2-8}          & \textbf{VisionAD (ours)} & -     & \textbf{74.7\footnotesize{$\pm$1.2}} & \textbf{70.1\footnotesize{$\pm$1.7}}  & \textbf{96.4\footnotesize{$\pm$0.3}} & \textbf{86.5\footnotesize{$\pm$1.1}}  & \textbf{81.9} \\
    \midrule    
    \multirow{7}{*}{4-shot} & SPADE~\cite{cohen2020sub} &  arXiv2020 & 50.8  & 45.8   & 59.5  & 19.2    & 43.8 \\
    & PaDiM~\cite{defard2021padim} &  ICPR2020 & 60.3  & 53.5   & 90.9  & 67.6    & 68.1 \\
    & PatchCore~\cite{roth2022towards} & CVPR2022 & 66.0 & 62.2  & 92.9  & 68.6  & 72.4 \\
    & WinCLIP~\cite{jeong2023winclip} & CVPR2023 & 73.0  & 61.8   & 93.8  & 76.4   & 76.2  \\
    & PromptAD~\cite{li2024promptad} & CVPR2024 & 59.7   & 43.5  & 86.9  & 61.9  & 63.0 \\ 
    & INP-Former~\cite{luo2025exploring} & CVPR2025 & \underline{76.7}   & \underline{72.3}  & \underline{97.3}  & \underline{89.0}  & \underline{83.8} \\
\cmidrule{2-8}          & \textbf{VisionAD (ours)} & -     & \textbf{79.1\footnotesize{$\pm$2.1}} & \textbf{74.7\footnotesize{$\pm$2.3}}  & \textbf{97.3\footnotesize{$\pm$0.3}} & \textbf{88.9\footnotesize{$\pm$1.4}}  & \textbf{85.0} \\
    \bottomrule
    \end{tabular}%
    }
  \label{realiad}
\end{table}

\begin{table}[t]
  \centering
    \caption{Performance comparisons under  full-shot multi-class(one-for-all) setting. Results below our 1-shot are marked in \textcolor{red}{red}, and those below our 8-shot are marked in \textcolor{blue}{blue}.}
    \setlength{\tabcolsep}{0.5mm}
    \resizebox{\linewidth}{!}{
    \begin{tabular}{lcccccc}
    \toprule
    \multirow{2}{*}{Method} & \multirow{2}{*}{Public} & \multirow{2}{*}{Setup} & \multicolumn{2}{c}{MVTec} & \multicolumn{2}{c}{VisA} \\
    \cmidrule(r){4-5}   \cmidrule(r){6-7}
    &    &    & AUROC & pAUROC & AUROC & pAUROC \\
    \midrule
    \multirow{4}{*}{VisionAD (Ours)}
        & \multirow{4}{*}{-}    & 1-shot & 97.4\footnotesize{$\pm$0.4}  & 96.2\footnotesize{$\pm$0.2}   & 94.8\footnotesize{$\pm$1.0}  & 97.6\footnotesize{$\pm$0.3} \\
        &                       & 2-shot & 98.1\footnotesize{$\pm$0.3}  & 96.6\footnotesize{$\pm$0.1}   & 95.0\footnotesize{$\pm$0.3}  & 97.7\footnotesize{$\pm$0.0} \\
        &                       & 4-shot & 98.6\footnotesize{$\pm$0.1}  & 96.9\footnotesize{$\pm$0.1}   & 95.7\footnotesize{$\pm$0.3}  & 98.0\footnotesize{$\pm$0.0} \\
        &                       & 8-shot & 98.9\footnotesize{$\pm$0.2}  & 97.1\footnotesize{$\pm$0.1}   & 96.3\footnotesize{$\pm$0.3}  & 98.2\footnotesize{$\pm$0.0} \\
    \midrule    
    UniAD~\cite{you2022unified} &  NeurIPS2022 & \multirow{7}{*}{full-shot} & \color[HTML]{FE0000}96.5  & \color[HTML]{3166FF}96.8   & \color[HTML]{FE0000}91.9  & 98.6 \\
    SimpleNet~\cite{liu2023simplenet} &  CVPR2023 & & \color[HTML]{FE0000}95.3  & \color[HTML]{3166FF}96.9   & \color[HTML]{FE0000}87.2  & \color[HTML]{FE0000}96.8 \\  
    DeSTSeg~\cite{zhang2023destseg} &  CVPR2023 &   & \color[HTML]{FE0000}89.2  & \color[HTML]{FE0000}93.1   & \color[HTML]{FE0000}88.9  & \color[HTML]{FE0000}96.1 \\    
    DiAD~\cite{he2024diffusion} &  AAAI2024 &  & \color[HTML]{FE0000}97.2  & \color[HTML]{3166FF}96.8   & \color[HTML]{FE0000}86.8  & \color[HTML]{FE0000}96.0 \\
    MambaAD~\cite{he2024mambaad} &  NeurIPS2024 & & \color[HTML]{3166FF}98.6  & 97.7   & \color[HTML]{FE0000}94.3  & 98.5 \\    
    ViTAD~\cite{zhang2023exploring} &  CVIU2025 & & \color[HTML]{3166FF}98.3  & 97.7   & \color[HTML]{FE0000}90.5  & \color[HTML]{3166FF}98.2 \\
    Dinomaly~\cite{guo2024dinomaly} &  CVPR2025 & & 99.6  & 98.4   & 98.7  & 98.7 \\    
    \bottomrule
    \end{tabular}%
    }
  \label{full-shot}
\end{table}

\begin{table}[!t]
  \centering
  \scriptsize
  \caption{Scaling of ViT model sizes on MVTec and VisA under 1-shot setting (\%). Im/s (Troughoutput, image per second) is measured on NVIDIA RTX4090 with batch size=8. \dag:default.}
   \setlength{\tabcolsep}{0.5mm}
   \resizebox{0.95\linewidth}{!}{
    \begin{tabular}{cccccccc}
    \toprule
    \multirow{2}{*}{Arch.} & \multirow{2}{*}{Params} & \multirow{2}{*}{MACs} & \multirow{2}{*}{Im/s} & \multicolumn{2}{c}{MVTec} & \multicolumn{2}{c}{VisA} \\
    \cmidrule(r){5-6} \cmidrule(l){7-8} 
    & & & & \multicolumn{1}{c}{AUROC} & \multicolumn{1}{c}{pAUROC} & \multicolumn{1}{c}{AUROC} & \multicolumn{1}{c}{pAUROC} \\ \midrule
    {ViT-Small} & 18.0M & 14.2G & 52.0  & 96.3 & 95.6  & 91.3  & 96.8 \\
    \midrule  
    ViT-Base    & 71.3M & 56.2G & 26.6  & \underline{97.6} & \underline{96.0}  & \underline{92.7}   & \underline{97.2} \\
    \midrule 
    ViT-Large\dag   & 239.9M & 189.2G & 10.1  & \textbf{97.7} & \textbf{96.2}  & \textbf{93.7} & \textbf{97.3} \\
    \bottomrule
    \end{tabular}
    }
\label{tab:arch_scale}
\end{table}

\begin{table}[!t]
  \centering
    \tiny
  \caption{Ablation study under 1-shot setting (\%). AF: Advanced Foundation. FF: Feature Fusion. SA: Support Augmentation. PMVT: Pseudo Multi-View Transformation. CIMB: Category-Indexed Memory Bank. CS: Class-Separated Setting. MC: Multi-Class Setting.}
  \setlength{\tabcolsep}{0.5mm}
   \resizebox{0.9\linewidth}{!}{
    \begin{tabular}{cccccccccc}
        \toprule
        \multirow{2}[2]{*}{Setting} & \multirow{2}[2]{*}{AF} & \multirow{2}[2]{*}{FF} & \multirow{2}[2]{*}{SA} & \multirow{2}[2]{*}{PMVT} & \multirow{2}[2]{*}{CIMB} & \multicolumn{2}{c}{MVTec} & \multicolumn{2}{c}{VisA} \\
        \cmidrule(r){7-8} \cmidrule(l){9-10} 
        & & & & & & \multicolumn{1}{c}{AUROC} & \multicolumn{1}{c}{pAUROC} & \multicolumn{1}{c}{AUROC} & \multicolumn{1}{c}{pAUROC} \\ \midrule
        \multirow{5}{*}{CS}    &    &    &    &    &    & 89.00  & 95.17 & 78.32 & 95.41 \\
        & \checkmark  &    &    &    &    & 92.83 & 94.44 & 80.50  & 93.57 \\
        & \checkmark  & \checkmark  &    &    &    & 94.96 & 95.50 & 86.81  & 96.24 \\
        & \checkmark  & \checkmark    & \checkmark  &    &    & \underline{97.07} & \underline{95.57} & \underline{90.17}  & \underline{96.36} \\
        & \checkmark  & \checkmark    & \checkmark  & \checkmark  &    & \textbf{97.60} & \textbf{96.01} & \textbf{92.66}  & \textbf{97.22} \\
        \midrule
        \multirow{2}{*}{MC} & \checkmark  & \checkmark    & \checkmark  & \checkmark  &    & 97.06 & 95.33  & 92.49    & \textbf{97.26}    \\
        & \checkmark  & \checkmark    & \checkmark  & \checkmark  & \checkmark    & \textbf{97.60} & \textbf{96.01} & \textbf{92.66}  & 97.22 \\
        \bottomrule
    \end{tabular}
    }
\label{tab:main_ablation}
\end{table}

\begin{table}[!t]
  \centering
  \scriptsize
  \caption{Performance using different support augmentations, view transformations under 1-shot setting.}
   \setlength{\tabcolsep}{0.8mm}
   \resizebox{\linewidth}{!}{
    \begin{tabular}{ccccc}
    \toprule
    \multirow{2}{*}{Aug.} & \multicolumn{2}{c}{MVTec} & \multicolumn{2}{c}{VisA} \\
    \cmidrule(r){2-3} \cmidrule(l){4-5} 
    & \multicolumn{1}{c}{AUROC} & \multicolumn{1}{c}{pAUROC} & \multicolumn{1}{c}{AUROC} & \multicolumn{1}{c}{pAUROC} \\ \midrule
    Baseline & 95.0 & 95.5  & 86.8  & 96.2 \\
    \midrule
    Baseline+Rot.\dag & 97.0 & 95.6  & 90.2  & 96.4 \\
    Baseline+Rot.+flip & 97.1 & 95.6  & 90.2  & 96.4 \\
    Baseline+Rot.+flip+affine & 97.1 & 95.7  & 90.1  & 96.4 \\
    \midrule
    Baseline+PosClamp & 95.5 & 95.9  & 88.8  & 97.0 \\
    Baseline+NegClamp & 94.8 & 95.4  & 84.2  & 96.0 \\
    Baseline+XFlip & 95.0 & 95.6  & 87.0  & 96.4 \\
    Baseline+YFlip & 95.3 & 95.7  & 87.1  & 96.4 \\
    Baseline+RBSwap & 95.0 & 95.5  & 86.7  & 96.3 \\
    Baseline+PosClamp+YFlip\dag & 95.5 & 96.0  & 89.2  & 97.1 \\
    \midrule
    Baseline+Rot.+PosClamp+YFlip & \textbf{97.6} & \textbf{96.0}  & \textbf{92.7}  & \textbf{97.2} \\
    \bottomrule
    \end{tabular}
    }
\label{tab:aug}
\end{table}

\begin{table}[!t]
  \centering
  \scriptsize
  \caption{Performance using different Fuser variants under 1-shot setting (without SA and PMVT). The numbers represent the usage of the corresponding layer features, where the variants, separated by commas, indicate the comparison and subsequent fusion to obtain the anomaly map.}
   \setlength{\tabcolsep}{1.3mm}
   \resizebox{0.9\linewidth}{!}{
    \begin{tabular}{ccccc}
    \toprule
    \multirow{2}{*}{Fuser Variant} & \multicolumn{2}{c}{MVTec} & \multicolumn{2}{c}{VisA} \\
    \cmidrule(r){2-3} \cmidrule(l){4-5} 
    & \multicolumn{1}{c}{AUROC} & \multicolumn{1}{c}{pAUROC} & \multicolumn{1}{c}{AUROC} & \multicolumn{1}{c}{pAUROC} \\ \midrule
    12(last layer) & 92.8 & 94.4  & 80.5  & 93.6 \\
    3,6,9,12 & 93.8 & \textbf{96.1}  & \textbf{87.4}  & \underline{96.0} \\
    345678910\dag & \textbf{95.0} & 95.5  & 86.8  & \textbf{96.2} \\
    3456,78910 & \underline{94.4} & \underline{95.9}  & \underline{87.2}  & 95.8 \\
    \bottomrule
    \end{tabular}
    }
\label{tab:fuser_variants}
\end{table}
\subsection{Comparisons with State-of-the-Arts}
Table~\ref{mvtec-visa} and Table~\ref{realiad} demonstrate the comparison results of VisionAD (ViT-Large) with existing few-shot anomaly detection methods~\cite{roth2022towards,jeong2023winclip,chen2023zero,gu2024anomalygpt,li2024promptad,tao2024kernel,lvone,cohen2020sub,defard2021padim,luo2025exploring}.
%
VisionAD shows significant improvements over all methods across the MVTec, VisA, and Real-IAD datasets in almost all metrics.
On the widely used MVTec-AD dataset, VisionAD achieves AUROC/AUPR/PRO performance of 97.4/99.0/92.5 (\%), outperforming the previous state-of-the-art (SoTA) methods by 1.6/0.9/1.7, respectively, in the 1-shot setting compared to the suboptimal method results.
On the popular VisA dataset, VisionAD achieves AUROC/AUPR/pAUROC/PRO performance of 94.8/95.0/97.6/91.6, surpassing previous SoTAs by 3.2/1.7/0.6/1.6.
On the Real-IAD dataset, which contains 30 classes with 5 camera views each, VisionAD produces AUROC/AUPR/pAUROC/PRO performance of 70.8/66.7/95.7/84.9, outperforming previous SoTAs by 1.4/3.6/0.8/3.1, demonstrating its scalability to highly complex scenarios.
Averaging the results across all metrics, VisionAD shows the best performance.
Furthermore, VisionAD achieves competitive results in the pAUROC metric on MVTec and state-of-the-art performance in the 2-shot and 4-shot settings.
The quantitative results of VisionAD in anomaly localization highlight its significant advantages in performance under different numbers of support images.

\subsection{Visualization Results}
\begin{figure}[t]
  \centering
  \includegraphics[width=\linewidth]{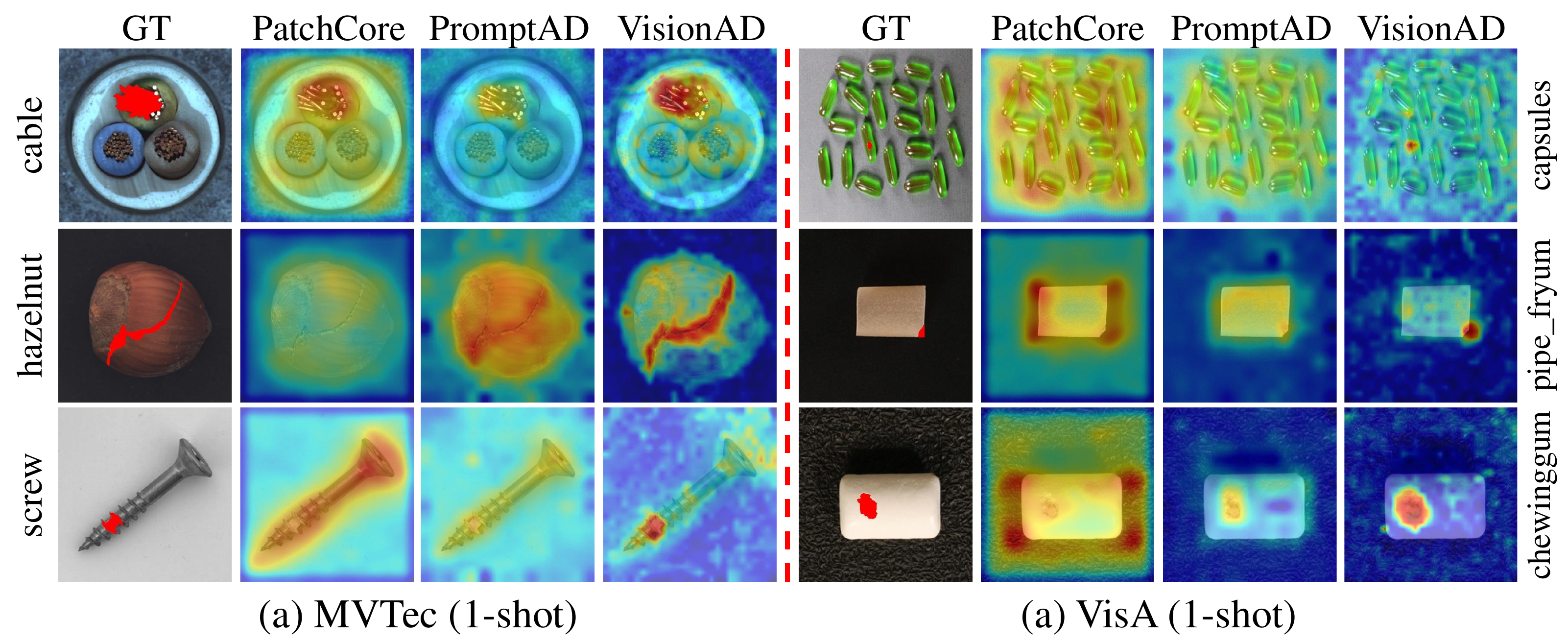}
  \caption{Qualitative comparison results of 1-shot pixel-level anomaly detection on MVTec~\cite{bergmann2019mvtec} and VisA~\cite{zou2022spot}.}
  \label{fig:qualitative}
\end{figure}

Fig.~\ref{fig:qualitative} shows the partial visualization results of VisionAD on MVTec and VisA dataset. Compared to PatchCore~\cite{roth2022towards} and PromptAD~\cite{li2024promptad}, VisionAD produces more distinctive and precise anomaly maps in the 1-shot setting. Additionally, VisionAD is capable of accurately locating subtle anomalies that are often overlooked. This robustness and reliability under few-shot settings underscore the potential practical applicability of VisionAD in real-world scenarios. For more qualitative results refer to Appendix~\ref{sec:visual}.

\subsection{Compared With Multi-class Full-Shot Methods}
In Table~\ref{full-shot}, we present the comparison results of VisionAD under few-shot settings with several prior works evaluated under the full-shot multi-class setting on the MVTec and VisA datasets. It can be observed that, compared to some methods under the full-shot setting, VisionAD achieves better image-level results and competitive pixel-level results. This fully demonstrates the strong ability of VisionAD in few-shot settings. Additionally, VisionAD outperforms early full-shot AD methods such as UniAD~\cite{you2022unified}, SimpleNet~\cite{liu2023simplenet}, DeSTSeg~\cite{zhang2023destseg}, and DiAD~\cite{he2024diffusion}, although there remains a gap between VisionAD and the latest full-shot one-for-all methods, including MambaAD~\cite{he2024mambaad} and Dinomaly~\cite{guo2024dinomaly}. This highlights the effectiveness of our method in leveraging limited data.

\subsection{Ablation Study}
\label{ablation}
\textbf{Overall Ablation.} We conduct experiments to verify the effectiveness of different modules: Advanced Foundation (AF), Feature Fusion (FF), Support Augmentation (SA), Pseudo Multi-View Transformation (PMVT), and Category-Indexed Memory Bank (CIMB). Image-level and pixel-level results on MVTec-AD and VisA under the 1-shot setting are shown in Table~\ref{tab:main_ablation}. In the first row, we provide a baseline model without incorporating any of the proposed modules, using the common DINO, the last layer feature, no augmentation, and no category-indexed memory bank. Each module contributes to the superior anomaly classification performance of VisionAD. A slight improvement is observed in anomaly segmentation due to the subtle anomaly areas. The combined use of all modules helps the model search more accurately. When applying VisionAD to multi-class FSAD scenarios, the use of the Category-Indexed Memory Bank does not result in any performance degradation, as the semantic differences in images are often quite clear. More ablation and detailed experiment results refer to Appendix~\ref{sec:ablation} and \ref{sec:perclass}.

\begin{figure}[t]
  \centering
  \includegraphics[width=\linewidth]{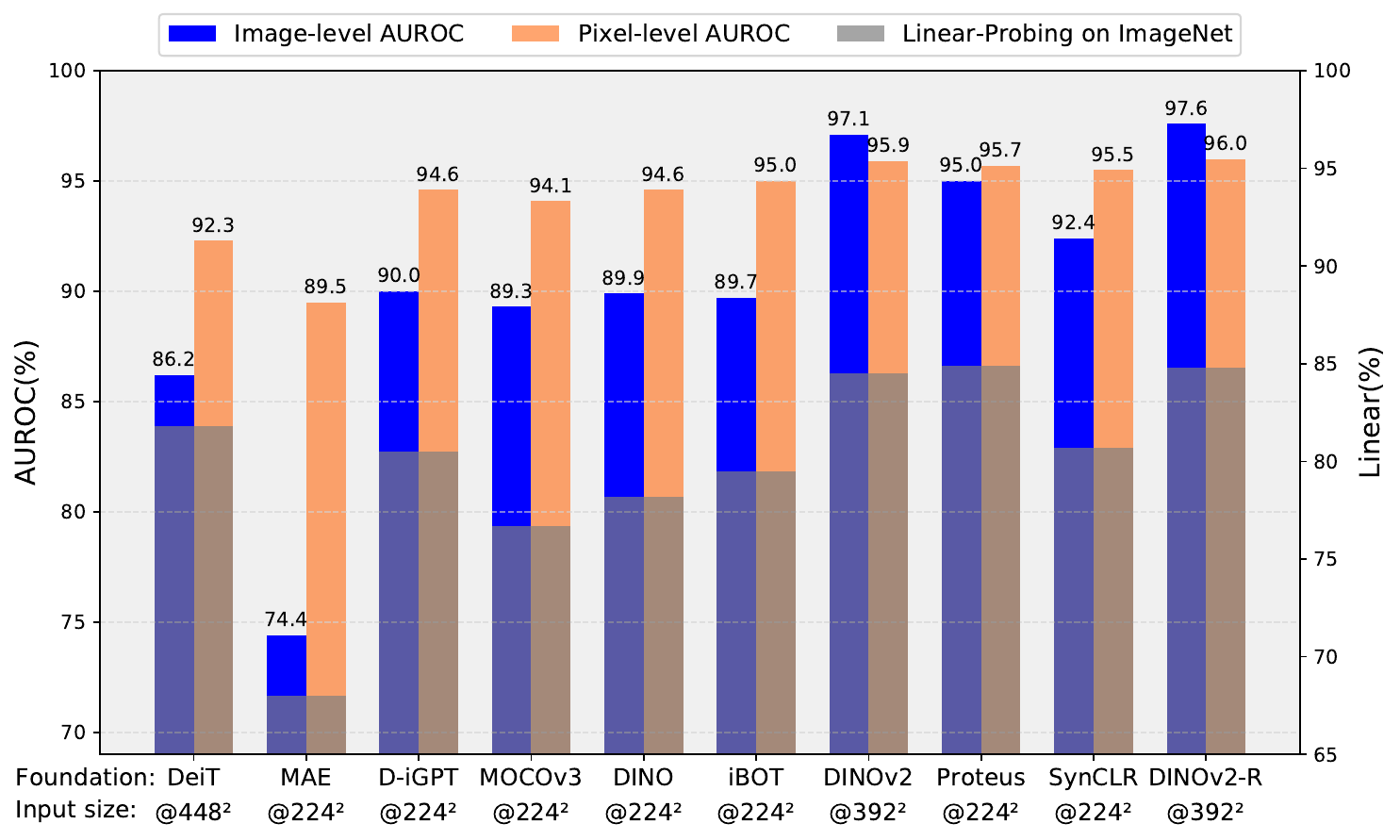}
  \caption{Image-level/pixel-level results on MVTec~\cite{bergmann2019mvtec} in the 1-shot setting using  various ViT-Base foundations, and their linear-probing accuracy on ImageNet.}
  \label{Change_backbone}
\end{figure}

\textbf{ViT Foundations.}
 We conduct extensive experiments to explore the impact of diverse pre-trained ViT-Base foundations, including DeiT \cite{touvron2021training}, MAE~\cite{he2022masked}, D-iGPT~\cite{ren2023rejuvenating}, MOCOv3~\cite{chen2021empirical}, DINO~\cite{caron2021emerging}, iBot~\cite{zhou2021ibot}, DINOv2~\cite{oquab2023dinov2}, Proteus~\cite{zhang2025accessing}, SynCLR~\cite{tian2024learning}, and DINOv2-R~\cite{darcet2023vision}.
As shown in Figure~\ref{Change_backbone}, compared to the less affected anomaly segmentation, anomaly classification performance shows a more pronounced difference. However, some foundation models can produce competitive results, with image-level/pixel-level AUROC exceeding 90\%/95\%.  
MAE has exceptionally low AUROC, which has been reported to be less effective across various unsupervised tasks (e.g., kNN and linear-probing) without fine-tuning~\cite{oquab2023dinov2}.  
The optimal input size varies because these foundations are pre-trained on different resolutions.
Interestingly, we found that anomaly detection performance is strongly correlated with the accuracy of ImageNet linear-probing (freezing the backbone and only tuning the linear classifier) of the foundation model, suggesting that future improvements could be achieved by simply adopting a more advanced foundation model.

Additionally, we found that the performance of the proposed VisionAD benefits from scaling, as shown in Table~\ref{tab:arch_scale}. ViT-Large further boosts VisionAD to an unprecedented level. This scalability allows users to select an appropriate model size based on the computational resources available in their specific scenario.
%

\textbf{Support Augmentation and Pseudo Multi-View Transformation.}
To validate the effectiveness of our support Augmentation and pseudo multi-view transformation design, we conducted an ablation study using different support augmentations and view transformations. The results are presented in Table~\ref{tab:aug}. The second section shows the performance of gradually adding rotation, flip, and affine transformations to support images. A significant performance improvement occurs when rotating support images, demonstrating that simple augmentation strategies allow the feature memory bank to store more useful and diverse features.
The third section shows the performance of applying PositiveClamp, NegativeClamp, XFlip, YFlip, R-B channel swap to query and support images. The performance improvement based on some view transformations indicates that pseudo multi-view enhances the accuracy and robustness of the predictions.

\textbf{Feature Fusion Variants.}
As shown in Table~\ref{tab:fuser_variants}, we conducted an ablation study to explore the impact of different layer feature fusion on the results. We first aggregate multi-layer feature similarity (Fig.~\ref{feature_fusion} (a) and the first row in Table~\ref{tab:fuser_variants}). Then, we employ a single feature group (Fig.~\ref{feature_fusion} (c) and the third row in Table~\ref{tab:fuser_variants}), and two feature groups (Fig.~\ref{feature_fusion} (d) and the fourth row in Table~\ref{tab:fuser_variants}). The results in the second row show the use of the last layer feature for search. When adopting the structure in Fig.~\ref{feature_fusion} (c) and (d), a significant improvement in AUROC is observed on MVTec, with almost identical performance on VisA. This indicates that our strategy leverages richer, more diversified features for search and comparison, enabling the detection of subtle local anomalies.

\section{Conclusion}
In this paper, we propose a training-free, vision-guided and minimalistic FSAD framework termed VisionAD which is proposed to address sophisticated prompt engineering and extensive manual tuning of existing methods. We present four key elements in VisionAD, i.e., Foundation Transformer, Dual Augmentation Strategies, Multi-layer Feature Integration, and a class-aware visual memory bank, that can boost the performance under the challenging few-shot setting without fancy modules and tricks. Extensive experiments on MVTec AD, VisA, and Real-IAD demonstrate our superiority over previous state-of-the-art approaches under FSAD setting.
\end{sloppypar}



\bibliographystyle{ACM-Reference-Format}
\bibliography{samples/main}

\appendix
\clearpage
\begin{center}
\textbf{SUPPLEMENTARY}
\end{center}

\setcounter{table}{0}
\renewcommand{\thetable}{A\arabic{table}}

\setcounter{figure}{0}
\renewcommand{\thefigure}{A\arabic{figure}}

\setcounter{section}{0} 
\renewcommand{\thesection}{\Alph{section}} 

\section{Additional Ablation and Experiment}
 \label{sec:ablation}

\textbf{Pre-Trained Foundations.} The representation quality of the frozen backbone Transformer is of great significance to FSAD. We conduct extensive experiments to probe the impact of different pre-training methods, including supervised learning and self-supervised learning. DeiT \cite{touvron2021training} is trained on ImageNet\cite{deng2009imagenet} in a supervised manner by distilling CNNs. Proteus~\cite{zhang2025accessing} distills foundation models into smaller equivalents on ImageNet-1K. SynCLR~\cite{tian2024learning} learns visual representations exclusively from synthetic images and synthetic captions, without relying on any real data. MAE~\cite{he2022masked}, BEiTv2~\cite{peng2002beitv2}, and D-iGPT~\cite{ren2023rejuvenating} are based on masked image modeling (MIM). MOCOv3~\cite{chen2021empirical} is based on contrastive learning (CL), pulling the representations of the similar images and pushing those of different images. DINO~\cite{caron2021emerging} is based on positive-pair contrastive learning, which is also referred to as self-distillation. iBot~\cite{zhou2021ibot} and DINOv2~\cite{oquab2023dinov2} combine MIM and CL strategies, marking the SoTA of self-supervised foundation models. DINOv2-R \cite{darcet2023vision} is a variation of DINOv2 that employs 4 extra register tokens. 
In addition, we evaluate the latest pre-training of large vision encoders, AIMv2, a family of open vision models pre-trained to autoregressively generate both image patches and text tokens.

It is worth noting that most models are pre-trained at a resolution of $224\times224$, except for DINOv2~\cite{oquab2023dinov2} and DINOv2-R~\cite{darcet2023vision}, which include an additional high-resolution training phase at $392\times392$. AIMv2 is pre-trained at $448\times448$ and $336\times336$ resolutions.
Directly applying pre-trained weights to a different resolution in FSAD without fine-tuning can degrade performance, as shown in Table~\ref{tab:pretrain_model}. Rather than significantly affecting pixel-level performance, the resolution mismatch has a more noticeable impact on image-level results. Proteus, DINOv2, and DINOv2-R, which are CL+MIM hybrid models, demonstrate better utilization of high-resolution inputs, confirming that strong backbone models can extract universal and discriminative features for FSAD. For example, the larger model AIMv2 also demonstrates performance that surpasses most other methods.
In contrast, using MAE leads to the worst performance, which can be attributed to weak semantic representations induced by its pretraining strategy.

\textbf{Input Size.} To evaluate the impact of input resolution on VisionAD's performance, we report its results under different resolutions, as shown in Table~\ref{tab:input_scale_}. Following PatchCore~\cite{roth2022towards}, we adopt center-crop preprocessing to reduce the influence of background regions. Experimental results demonstrate that VisionAD benefits from higher input resolutions for both anomaly detection and localization, while still achieving state-of-the-art performance even with smaller images. For fairness, all localization experiments are evaluated at a fixed resolution of $256\times256$.

\textbf{Scalability on MVTec and VisA under 1-/2-/4-shot setting.}
We demonstrate the performance of different ViT model sizes on MVTec and VisA under 1-/2-/4-shot setting in Table~\ref{tab:arch_scale_all}. The results show that larger models and more support images can lead to better performance. However, this also increases computational cost and slows down inference speed.

\section{Results Per-Category}
\label{sec:perclass}
For future research, we report the per-class results of MVTec-AD~\cite{bergmann2019mvtec}, VisA~\cite{zou2022spot}, and Real-IAD~\cite{wang2024real}. The performance of compared methods is drawn from KAG-prompt~\cite{tao2024kernel} and One-for-All~\cite{lvone}. The results of the 1-/2-/4-shot setting on MVTec-AD are presented in Table~\ref{tab:mvtec_1-shot}, Table~\ref{tab:mvtec-2-shot} and Table~\ref{tab:mvtec-4-shot}, respectively. The results of the 1-/2-/4-shot setting on VisA are presented in Table~\ref{tab:visa-1-shot}, Table~\ref{tab:visa-2-shot} and Table~\ref{tab:visa-4-shot}, respectively. The results of the 1-/2-/4-shot setting on Real-IAD are presented in Table~\ref{tab:real-iad}.

\section{Qualitative Visualization}
\label{sec:visual}

\begin{figure*}[!ht]
    \centerline{\includegraphics[width=0.9\textwidth]{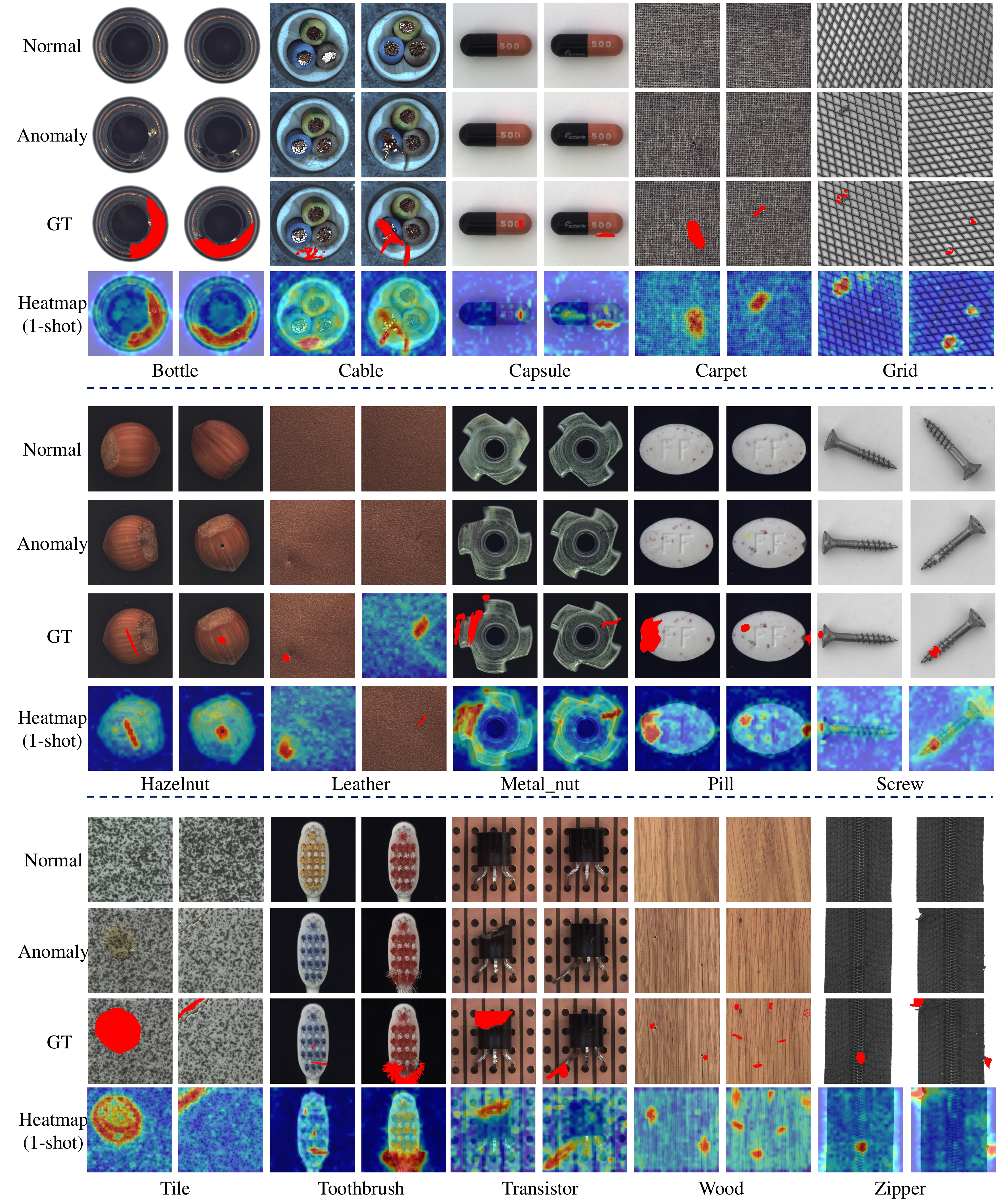}}
    \caption{Additional qualitative 1-shot results of our method on MVTec. All samples are randomly selected.}
\label{fulu_mvtec}
\end{figure*}

\begin{figure*}[!ht]
    \centerline{\includegraphics[width=0.9\textwidth]{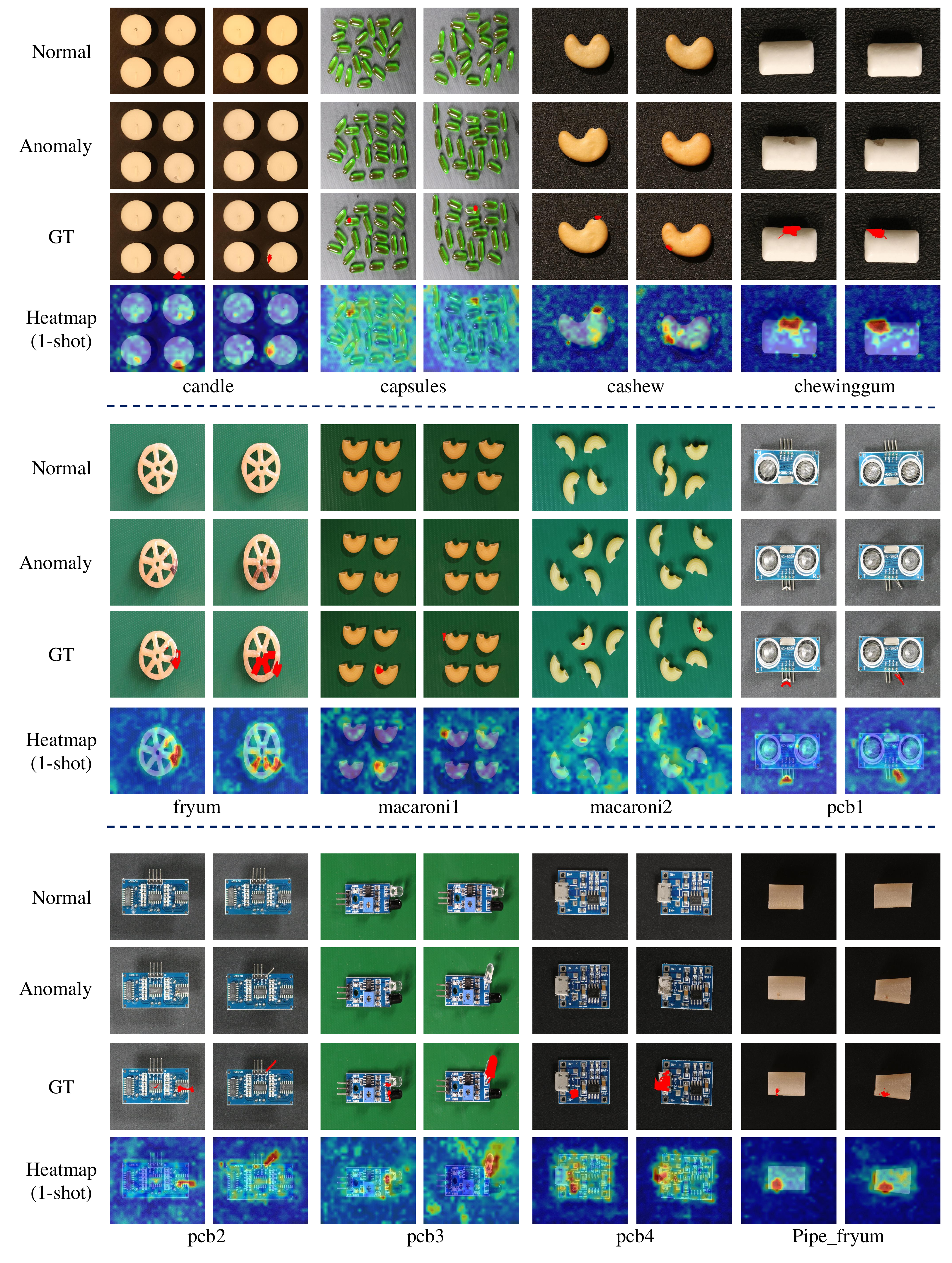}}
    \caption{Additional qualitative 1-shot results of our method on VisA. All samples are randomly selected.}
\label{fulu_visa}
\end{figure*}

\begin{figure*}[!h]
    \centerline{\includegraphics[width=0.72\textwidth]{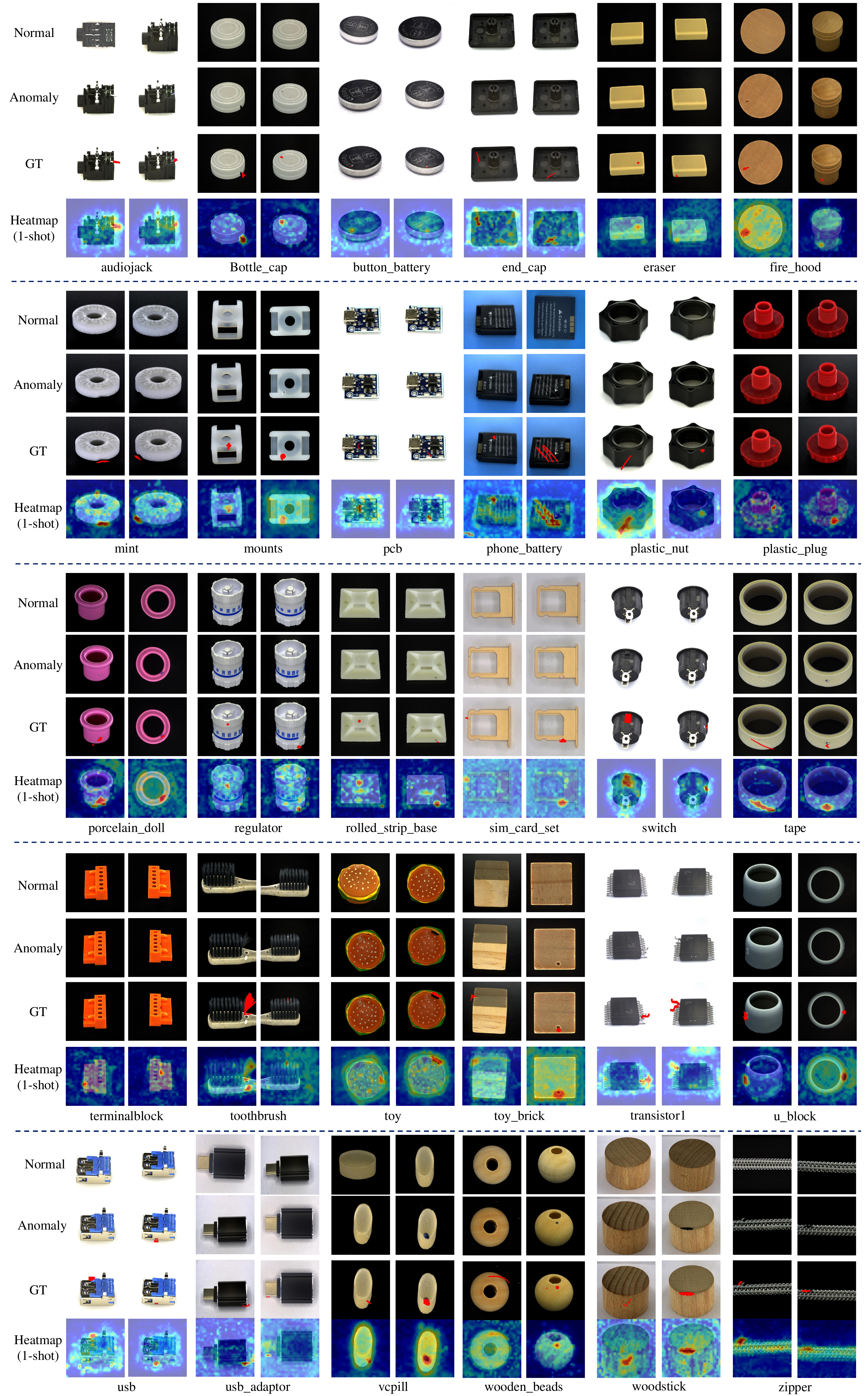}}
    \caption{Additional qualitative 1-shot results of our method on Real-IAD. All samples are randomly selected.}
\label{visul_realiad}
\end{figure*}

We visualize the output anomaly maps of VisionAD on MVTec-AD, VisA, and Real-IAD, as shown in Figure~\ref{fulu_mvtec}, Figure~\ref{fulu_visa} and Figure~\ref{visul_realiad}. It is noted that all visualized samples are randomly chosen without artificial selection. It shown that VisionAD has achieved anomaly localization results that closely match their ground truth. It not only effectively identifies larger anomalous regions but also detects subtle anomalies that are often overlooked. This demonstrates the exceptional anomaly localization capability of VisionAD. Such robustness and reliability under few-shot settings highlight the potential practical applicability of VisionAD in real-world scenarios. 

\section{Additional Dataset}

To further assess the applicability and generalization of the proposed CRR, we apply it to two real-world industrial tasks under both class-separated and multi-class settings, including the CSDD and HSS-IAD datasets. The CSDD~\cite{10502267} (Casting Surface Defect Detection) dataset aims to bridge the gap between real-world defect detection and the idealized conditions of existing datasets. It contains 12,647 high-resolution grayscale images with pixel-precise ground truth annotations for all defect samples. The HSS-IAD (Heterogeneous Same-Sort Industrial Anomaly Detection) dataset includes 8,580 images of metallic-like industrial parts with detailed anomaly annotations. The parts in this dataset exhibit diverse structures and appearances, with subtle defects that closely resemble the base materials.

In Table~\ref{fig:other_dataset}, the quantitative results on CSDD and HSS-IAD under different settings are presented. As the number of support images increases, anomaly detection and segmentation performance significantly improve.
Under the multi-class setting, VisionAD struggles to accurately distinguish among casting categories of HSS-IAD, as the casting surfaces share highly similar semantics. This leads to a slight drop in the mean performance on HSS-IAD. For CSDD, it is also challenging to accurately find the corresponding support set for each test image due to the semantic similarity among samples of the same class. Therefore, we do not report performance results for this setting.
Considering the complexity and high accuracy required in real-world scenarios, the performance of current FSAD methods still falls short of meeting practical industrial demands.

\section{Failure Cases Study}
\begin{figure}[htb]
    \centerline{\includegraphics[width=0.4\textwidth]{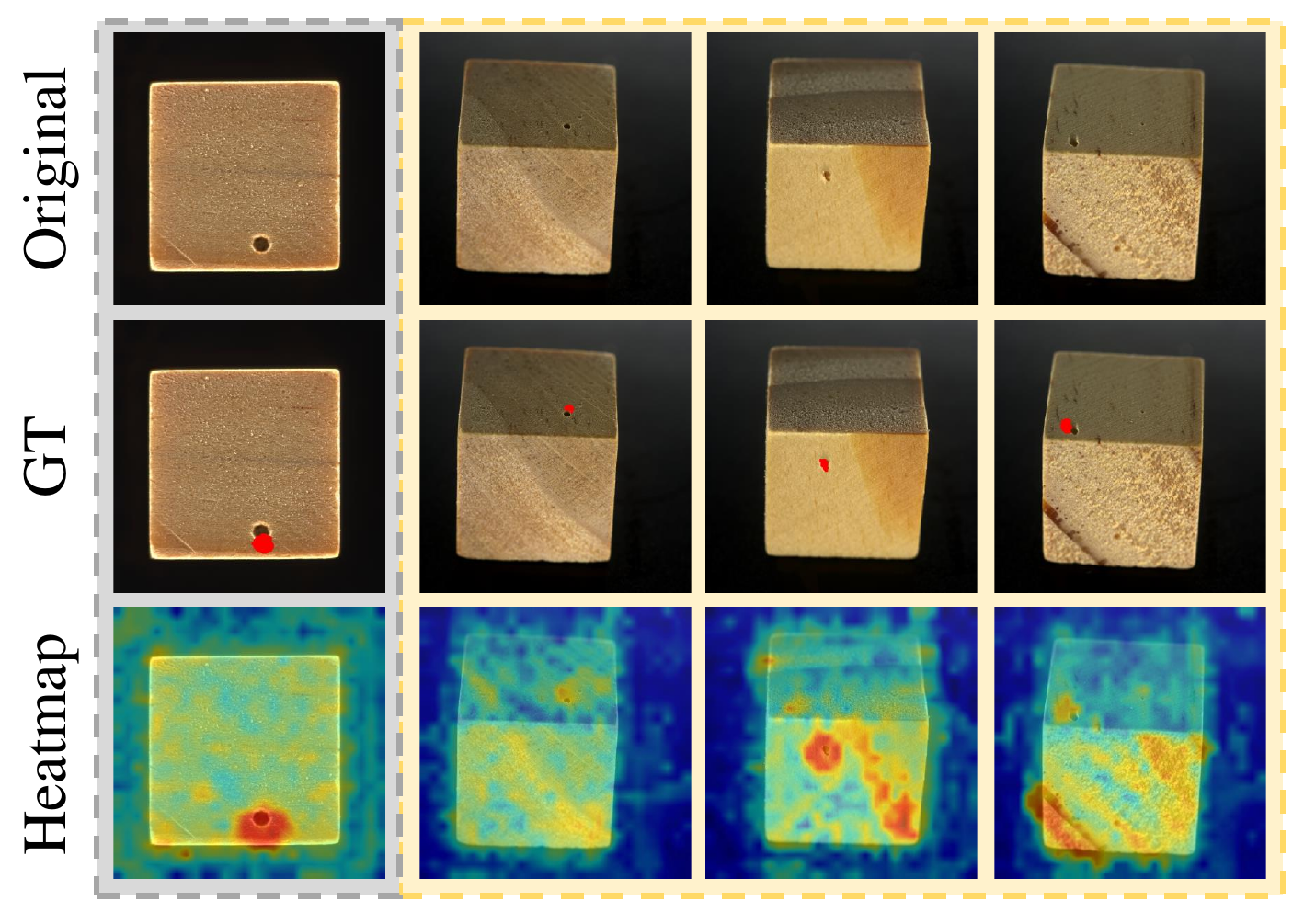}}
    \caption{Failure case on Real-IAD: The first column shows samples that share the same view with the supports, while the last three columns show the anomaly maps under different views.}
\label{realiad_failure_}
\end{figure}

To ensure a fair comparison with the INP on the Real-IAD dataset, we similarly selected only 1, 2, or 4 images per category as support samples. However, during the analysis of anomaly location (visualizations), we observed that VisionAD produced accurate results for images of the sample captured from similar views, whereas its performance noticeably degraded when the views varied significantly, as shown in Figure~\ref{realiad_failure_}. This observation confirms our initial hypothesis that the notion of 1-/2-/4-shot should fundamentally refer to the number of samples rather than the number of images. Therefore, future research should consider collecting at least five distinct views per sample to construct the support set.


\begin{table}[htb]
  \centering
  \scriptsize
  \caption{Scaling of ViT model sizes on MVTec and VisA (\%). Im/s (Troughoutput, image per second) is measured on NVIDIA RTX4090 with batch size=16. \dag:default.}
   \setlength{\tabcolsep}{0.8mm}
   \resizebox{\linewidth}{!}{
    \begin{tabular}{ccccccccc}
    \toprule
    \multirow{2}{*}{Arch.} & \multirow{2}{*}{Params} & \multirow{2}{*}{MACs} & \multirow{2}{*}{Im/s} & \multirow{2}{*}{Setup} & \multicolumn{2}{c}{MVTec} & \multicolumn{2}{c}{VisA} \\
    \cmidrule(r){6-7} \cmidrule(l){8-9} 
    & & & & & \multicolumn{1}{c}{AUROC} & \multicolumn{1}{c}{pAUROC} & \multicolumn{1}{c}{AUROC} & \multicolumn{1}{c}{pAUROC} \\ \midrule
    \multirow{3}{*}{ViT-Small}
    & \multirow{3}{*}{18.0M} & \multirow{3}{*}{14.2G} & \multirow{3}{*}{52.0}  & 1-shot & 96.3 & 95.6  & 91.3  & 96.8 \\
    &                        &   &  & 2-shot & 96.9 & 96.1  & 92.4  & 97.1 \\
    &                        &   &  & 4-shot & 97.3 & 96.4  & 94.1  & 97.5 \\
    \midrule  
    \multirow{3}{*}{ViT-Base\dag}
    & \multirow{3}{*}{71.3M} & \multirow{3}{*}{56.2G} & \multirow{3}{*}{26.6}  & 1-shot & 97.6 & 96.0  & 92.7   & 97.2 \\
    &                        &   &  & 2-shot & 98.1 & 96.5  & 93.6  & 97.5 \\
    &                        &   &  & 4-shot & \underline{98.5} & \underline{96.7}  & \underline{95.1}  & \underline{97.9} \\
    \midrule 
    \multirow{3}{*}{ViT-Large}
    & \multirow{3}{*}{239.9M} & \multirow{3}{*}{189.2G} & \multirow{3}{*}{10.1}  & 1-shot & 97.7 & 96.2  & 93.7 & 97.3 \\
    &                        &   &  & 2-shot & 98.3 & 96.7  & 94.5  & 97.6 \\
    &                        &   &  & 4-shot & \textbf{98.7} & \textbf{96.9}  & \textbf{96.1}  & \textbf{98.0} \\
    \bottomrule
    \end{tabular}
    }
\label{tab:arch_scale_all}
\end{table}

\begin{table}[htb]
  \centering
  \scriptsize
  \caption{Scaling input size on MVTec-AD and VisA (\%). \dag: default.}
   \setlength{\tabcolsep}{0.8mm}
   \resizebox{\linewidth}{!}{
    \begin{tabular}{ccccccccc}
    \toprule
    \multirow{2}{*}{Input Size} & \multirow{2}{*}{MACs} & \multirow{2}{*}{Setup} & \multicolumn{2}{c}{MVTec} & \multicolumn{2}{c}{VisA} \\
    \cmidrule(r){4-5} \cmidrule(l){6-7} 
    & & & \multicolumn{1}{c}{AUROC} & \multicolumn{1}{c}{pAUROC} & \multicolumn{1}{c}{AUROC} & \multicolumn{1}{c}{pAUROC} \\ \midrule
    \multirow{3}{*}{R$256^2$-C$224^2$}
    & \multirow{3}{*}{18.6G}     & 1-shot & 95.5 & 95.4  & 88.6  & 96.1 \\
    &                        & 2-shot & 96.5 & 96.1  & 89.8  & 96.4 \\
    &                        & 4-shot & 97.0 & 96.3  & 91.2  & 97.0 \\
    \midrule
    \multirow{3}{*}{R$320^2$-C$280^2$}
    & \multirow{3}{*}{28.9G} & 1-shot & 97.2 & 95.7  & 90.9  & 96.9 \\
    &                        & 2-shot & 97.7 & 96.3  & 91.8  & 97.1 \\
    &                        & 4-shot & 98.1 & 96.5  & 93.4  & 97.6 \\
    \midrule
    \multirow{3}{*}{R$384^2$-C$336^2$}
    & \multirow{3}{*}{41.4G} & 1-shot & 97.5 & 95.9  & 91.5  & 97.2 \\
    &                        & 2-shot & 98.1 & 96.4  & 92.5  & 97.4 \\
    &                        & 4-shot & \underline{98.5} & \underline{96.6}  & 94.2  & \underline{97.8} \\
    \midrule    
    \multirow{3}{*}{R$448^2$-C$392^2$\dag}
    & \multirow{3}{*}{56.2G} & 1-shot & 97.6 & 96.0  & 92.7  & 97.2 \\
    &                        & 2-shot & 98.1 & 96.5  & 93.6  & 97.5 \\
    &                        & 4-shot & \underline{98.5} & \textbf{96.7}  & \underline{95.1}  & \textbf{97.9} \\
    \midrule 
    \multirow{3}{*}{R$512^2$-C$448^2$}
    & \multirow{3}{*}{73.4G} & 1-shot & 97.6 & 96.1  & 93.3  & 97.3 \\
    &                        & 2-shot & 98.3 & 96.5  & 94.1  & 97.6 \\
    &                        & 4-shot & \textbf{98.7} & \textbf{96.7}  & \textbf{95.9}  & \textbf{97.9} \\    
    \bottomrule
    \end{tabular}
    }
\label{tab:input_scale_}
\end{table}

\begin{table*}[hbt]
  \centering
    \caption{Performance on CSDD and HSS-IAD under multi-class and class-separated UAD setting (\%). CS: Class-Separated Setting. MC: Multi-Class Setting.}
    \setlength{\tabcolsep}{2.0mm}
    \resizebox{0.95\linewidth}{!}{
    \begin{tabular}{llcccccccc}
    \toprule
    \multirow{2}{*}{Setting} & \multirow{2}{*}{Setup} & \multicolumn{4}{c}{CSDD} & \multicolumn{4}{c}{HSS-IAD} \\
    \cmidrule(r){3-6}   \cmidrule(r){7-10}
    &    & AUROC & AUPR    & pAUROC    & PRO   & AUROC & AUPR    & pAUROC    & PRO \\
    \midrule
    \multirow{4}{*}{CS}
        & 1-shot & 48.5\footnotesize{$\pm$1.8}  & 57.1\footnotesize{$\pm$1.2}   & 69.9\footnotesize{$\pm$1.8}  & 27.5\footnotesize{$\pm$1.3}    & 68.8\footnotesize{$\pm$1.1}   & 74.1\footnotesize{$\pm$1.1}   & 85.4\footnotesize{$\pm$0.6}   & 65.8\footnotesize{$\pm$1.1} \\
        & 2-shot & 49.8\footnotesize{$\pm$2.2}  & 57.6\footnotesize{$\pm$1.9}   & 69.9\footnotesize{$\pm$1.9}  & 27.3\footnotesize{$\pm$2.1}    & 69.2\footnotesize{$\pm$0.9}   & 74.0\footnotesize{$\pm$0.6}   & 85.7\footnotesize{$\pm$0.2}   & 65.6\footnotesize{$\pm$0.8} \\
        & 4-shot & 50.3\footnotesize{$\pm$0.9}  & 57.4\footnotesize{$\pm$0.9}   & 73.0\footnotesize{$\pm$0.8}  & 30.9\footnotesize{$\pm$2.3}    & 69.6\footnotesize{$\pm$1.1}   & 73.7\footnotesize{$\pm$1.3}   & 86.7\footnotesize{$\pm$0.3}   & 66.9\footnotesize{$\pm$0.5} \\
        & 8-shot & 53.0\footnotesize{$\pm$1.7}  & 58.9\footnotesize{$\pm$1.2}   & 75.3\footnotesize{$\pm$0.8}  & 34.8\footnotesize{$\pm$2.1}    & 71.7\footnotesize{$\pm$1.7}   & 74.8\footnotesize{$\pm$1.2}   & 87.5\footnotesize{$\pm$0.5}   & 67.6\footnotesize{$\pm$1.2} \\
    \midrule
    
    \multirow{4}{*}{MC}
        & 1-shot & -  & -   & -  & -    & 67.7\footnotesize{$\pm$1.0}   & 73.7\footnotesize{$\pm$1.1}   & 84.2\footnotesize{$\pm$1.7}   & 63.3\footnotesize{$\pm$2.9} \\
        & 2-shot & -  & -   & -  & -    & 67.8\footnotesize{$\pm$1.4}   & 73.2\footnotesize{$\pm$1.3}   & 84.6\footnotesize{$\pm$1.4}   & 63.2\footnotesize{$\pm$1.9} \\
        & 4-shot & -  & -   & -  & -    & 67.8\footnotesize{$\pm$1.8}   & 72.8\footnotesize{$\pm$1.4}   & 85.3\footnotesize{$\pm$0.7}   & 64.5\footnotesize{$\pm$1.0} \\
        & 8-shot & -  & -   & -  & -    & 69.0\footnotesize{$\pm$2.0}   & 73.3\footnotesize{$\pm$1.4}   & 85.8\footnotesize{$\pm$0.9}   & 64.7\footnotesize{$\pm$2.0} \\

    \bottomrule
    \end{tabular}%
    }
  \label{fig:other_dataset}
\end{table*}

\begin{table*}[htbp]
  \centering
  \tiny
  \caption{Comparison between pre-trained ViT foundations under 1-shot setting, conducted on MVTec-AD and VisA (\%). All models except AIMv2 are ViT-Base. The patch size for AIMv2, DINOv2, and DINOv2-R is $14^2$, while for the others it is $16^2$. R$448^2$-C$392^2$ represents first resizing images to 448$\times$448, then center cropping to 392$\times$392.}
  \setlength{\tabcolsep}{0.8mm}
   \resizebox{\linewidth}{!}{
    \begin{tabular}{lcccccccccc}
    \toprule
    \multirow{2}[2]{*}{\makecell[c]{Pre-Train\\Backbone}} & \multirow{2}[2]{*}{Type} & \multirow{2}[2]{*}{\makecell[c]{Image\\Size}} & \multicolumn{4}{c}{MVTec} & \multicolumn{4}{c}{VisA} \\
    \cmidrule(r){4-7} \cmidrule(l){8-11} 
 & & & \multicolumn{1}{c}{AUROC} & \multicolumn{1}{c}{AUPR} & \multicolumn{1}{c}{pAUROC} & \multicolumn{1}{c}{PRO} & \multicolumn{1}{c}{AUROC} & \multicolumn{1}{c}{AUPR} & \multicolumn{1}{c}{pAUROC} & \multicolumn{1}{c}{PRO} \\
 \midrule
DeiT~\cite{touvron2021training} & Supervised & R$512^2$-C$448^2$   & 86.21 & 93.33 & 92.31  & 81.17 & 86.34  & 87.93 & 95.46 & 83.04 \\
Proteus~\cite{zhang2025accessing} & Supervised & R$448^2$-C$392^2$  & 95.89 & 98.21 & 96.11 & 90.41 & 89.23 & 90.04 & 96.37 & 82.46 \\
SynCLR~\cite{tian2024learning} & Supervised & R$448^2$-C$384^2$  & 82.18    & 96.69 & 95.57 & 88.04 & 87.92 & 89.40 & 96.58 & 85.78 \\
MAE~\cite{he2022masked} & MIM & R$512^2$-C$448^2$  & 72.75   & 86.95 & 86.79 & 70.11 & 70.31 & 73.45 &85.64  & 67.13 \\
D-iGPT~\cite{ren2023rejuvenating} & MIM & R$512^2$-C$448^2$  & 86.92 & 93.60 & 94.47 & 84.10 & 87.85 & 88.86 & 95.75 & 84.72 \\
MOCOv3~~\cite{chen2021empirical} & CL & R$512^2$-C$448^2$  & 88.10    & 94.03 & 94.71 & 85.48 & 85.45 & 86.94 & 96.07 & 84.25 \\
DINO~\cite{caron2021emerging} & CL & R$512^2$-C$448^2$  & 88.83  & 94.87 & 95.13 & 86.53 & 85.32 & 87.06 & 96.66 & 86.23 \\
iBOT~\cite{zhou2021ibot} & CL+MIM & R$512^2$-C$448^2$  & 88.66   & 94.76 & 95.46 & 86.83 & 85.18 & 87.16 & 96.70 & 85.79 \\
DINOv2~\cite{oquab2023dinov2} & CL+MIM & R$448^2$-C$392^2$  & 97.05  & 98.88 & 95.89 & 91.74 & 93.44 & 93.56 & 97.37 & 90.78 \\
DINOv2-R~\cite{darcet2023vision} & CL+MIM & R$448^2$-C$392^2$  & 97.60  & 99.11 & 96.01 & 91.86 & 92.66 & 92.74 & 97.22 & 90.93 \\
\midrule
DeiT\cite{touvron2021training} & Supervised & R$256^2$-C$224^2$   & 88.62    & 95.33    & 92.90 & 80.39 & 81.30 & 84.28 & 93.24 & 70.91 \\
Proteus~\cite{zhang2025accessing} & Supervised & R$256^2$-C$224^2$  & 94.95 & 97.95 & 95.72 & 88.88 & 89.23 & 90.04 & 96.37 & 82.46 \\
SynCLR~\cite{tian2024learning} & Supervised & R$256^2$-C$224^2$  & 92.44    & 96.87 & 95.51 & 87.82 & 84.70 & 86.67 & 94.92 & 75.43 \\
MAE~\cite{he2022masked} & MIM & R$256^2$-C$224^2$  & 74.39   & 87.64 & 89.45 & 70.36 & 64.77 & 69.53 & 85.43 & 55.20 \\
BEiTv2~\cite{peng2002beitv2} & MIM & R$256^2$-C$224^2$  & 93.26  & 97.30 & 94.50 & 86.17 & 86.04 & 86.67 & 93.20 & 70.96 \\
D-iGPT~\cite{ren2023rejuvenating} & MIM & R$256^2$-C$224^2$  & 89.95 & 95.59 & 94.58 & 84.95 & 86.82 & 88.93 & 94.67 & 78.19 \\
MOCOv3~\cite{chen2021empirical} & CL & R$256^2$-C$224^2$  & 89.29    & 95.81 & 94.14 & 83.15 & 85.45 & 86.94 & 96.07 & 84.25 \\
DINO~\cite{caron2021emerging} & CL & R$256^2$-C$224^2$  & 89.92  & 96.07 & 94.57 & 84.64 & 81.15 & 83.95 & 94.95 & 74.97 \\
iBOT~\cite{zhou2021ibot} & CL+MIM & R$256^2$-C$224^2$  & 89.73   & 96.02 & 94.97 & 84.98 & 81.48 & 84.22 & 94.96 & 75.09 \\
DINOv2~\cite{oquab2023dinov2} & CL+MIM & R$256^2$-C$224^2$  & 95.23  & 98.05 & 95.42 & 88.74 & 89.23 & 90.28 & 96.03 & 81.66 \\
DINOv2-R~\cite{darcet2023vision} & CL+MIM & R$256^2$-C$224^2$  & 95.53   & 98.23 & 95.41 & 88.96 & 88.55 & 89.80 & 96.12 & 82.66 \\
\midrule
AIMv2-large-patch14-336~\cite{fini2024multimodal}  & Supervised & R$384^2$-C$336^2$  &  93.99   & 97.32 & 94.53 & 88.99 & 90.32 & 91.29 & 96.50 & 88.31 \\
AIMv2-huge-patch14-336~\cite{fini2024multimodal}  & Supervised & R$384^2$-C$336^2$  & 93.72 & 97.00 & 94.17 & 88.49 & 90.59 & 91.93 & 96.72 & 88.68 \\
AIMv2-1B-patch14-336~\cite{fini2024multimodal}  & Supervised & R$384^2$-C$336^2$  & 94.33   & 97.36 & 93.92 & 88.67 & 91.71 & 92.89 & 96.87 & 89.50 \\
AIMv2-3B-patch14-336~\cite{fini2024multimodal}  & Supervised & R$384^2$-C$336^2$  & 95.32   & 97.98 & 93.90 & 88.09 & 92.17 & 93.40 & 96.70 & 89.04 \\
\midrule
AIMv2-large-patch14-448~\cite{fini2024multimodal}  & Supervised & R$512^2$-C$448^2$  & 93.89    & 97.21 & 94.04 & 88.97 & 92.62 & 93.19 & 96.56 & 90.21 \\
AIMv2-huge-patch14-448~\cite{fini2024multimodal}    & Supervised & R$512^2$-C$448^2$  & 93.52   & 96.86 & 93.51 & 88.08 & 92.68 & 93.84 & 96.64 & 90.11 \\
AIMv2-1B-patch14-448~\cite{fini2024multimodal}    & Supervised & R$512^2$-C$448^2$  & 94.25 & 97.24 & 93.51 & 88.58 & 93.38 & 94.27 & 96.76 & 90.41 \\
AIMv2-3B-patch14-448~\cite{fini2024multimodal}    & Supervised & R$512^2$-C$448^2$  & 95.04 & 97.67 & 93.06 & 87.42 & 93.80 & 94.39 & 96.60 & 90.46 \\
\bottomrule
\end{tabular}
}
\label{tab:pretrain_model}
\end{table*}


\begin{table*}[!h]
\centering
\caption{Subset-wise performance comparison results of the 1-shot setting for AUROC/AP/pAUROC/PRO on MVTecAD.}
\setlength{\tabcolsep}{0.5mm}
\resizebox{\linewidth}{!}{
\begin{tabular}{ccccccccccc}
\toprule
\multicolumn{2}{c}{Method~$\rightarrow$} & PatchCore~\cite{roth2022towards} & WinCLIP~\cite{jeong2023winclip} & APRIL-GAN~\cite{chen2023zero} & AnomalyGPT~\cite{gu2024anomalygpt} & PromptAD~\cite{li2024promptad} & KAG-prompt~\cite{tao2024kernel} & One-for-All~\cite{lvone} & \cellcolor{tab_ours}VisionAD \\
\cline{1-2}
\multicolumn{2}{c}{Category~$\downarrow$} & CVPR'22 & CVPR'23 & arXiv'23 & AAAI'24 & CVPR'24 & AAAI'25 & ICLR'25 & \cellcolor{tab_ours}Ours \\
\hline
\multicolumn{1}{c}{\multirow{10}[1]{*}{\begin{turn}{-90}Objects\end{turn}}} & \multicolumn{1}{c}{Bottle} & 99.4/99.8/97.9/93.5 & 98.2/99.4/97.5/91.2 & 99.9/100.0/98.7/96.5 & 98.1/98.7/97.6/94.4 & 99.8/99.8/99.6/93.6 & 98.7/99.6/97.5/93.9 & 99.7/99.9/98.4/94.9 & \cellcolor{tab_ours}97.3{\footnotesize{$\pm$0.3}}/88.3{\footnotesize{$\pm$0.1}}/96.6{\footnotesize{$\pm$0.0}}/93.3{\footnotesize{$\pm$0.1}}  \\

\multicolumn{1}{c}{} & \multicolumn{1}{c}{\cellcolor{tab_others}Cable} & \cellcolor{tab_others}88.8/93.8/95.5/84.7 & \cellcolor{tab_others}88.9/93.2/93.8/72.5 & \cellcolor{tab_others}92.8/97.8/96.4/92.7 & \cellcolor{tab_others}88.8/93.1/93.1/83.5 & \cellcolor{tab_others}94.2/95.5/98.4/87.3 & \cellcolor{tab_others}90.7/94.5/91.6/80.8  & \cellcolor{tab_others}92.8/96.1/94.9/86.7 & \cellcolor{tab_ours}93.1{\footnotesize{$\pm$0.7}}/96.5{\footnotesize{$\pm$0.3}}/93.4{\footnotesize{$\pm$0.4}}/85.4{\footnotesize{$\pm$1.1}}/ \\

\multicolumn{1}{c}{} & \multicolumn{1}{c}{Capsule} & 67.8/89.4/95.6/83.9 & 72.3/91.6/94.6/85.6 & 98.5/99.3/97.6/92.1 & 90.1/97.8/92.3/85.5 & 84.6/97.8/99.5/80.1 & 90.9/98.0/97.4/93.6 & 80.5/95.4/97.2/91.3 & \cellcolor{tab_ours}95.4{\footnotesize{$\pm$1.8}}/99.0{\footnotesize{$\pm$0.4}}/97.5{\footnotesize{$\pm$0.2}}/96.2{\footnotesize{$\pm$0.2}} \\
\bottomrule
\end{tabular}
}
\label{tab:mvtec_1-shot}%
\end{table*}%

\begin{table*}[!h]
\centering
\caption{Subset-wise performance comparison results of the 2-shot setting for AUROC/AP/pAUROC/PRO on MVTecAD.}
\setlength{\tabcolsep}{0.5mm}
\resizebox{\linewidth}{!}{
\begin{tabular}{ccccccccccc}
\toprule
\multicolumn{2}{c}{Method~$\rightarrow$} & PatchCore~\cite{roth2022towards} & WinCLIP~\cite{jeong2023winclip} & APRIL-GAN~\cite{chen2023zero} & AnomalyGPT~\cite{gu2024anomalygpt} & PromptAD~\cite{li2024promptad} & KAG-prompt~\cite{tao2024kernel} & One-for-All~\cite{lvone} & \cellcolor{tab_ours}VisionAD \\
\cline{1-2}
\multicolumn{2}{c}{Category~$\downarrow$} & CVPR'22 & CVPR'23 & arXiv'23 & AAAI'24 & CVPR'24 & AAAI'25 & ICLR'25 & \cellcolor{tab_ours}Ours \\
\hline
\multicolumn{1}{c}{\multirow{10}[1]{*}{\begin{turn}{-90}Objects\end{turn}}} & \multicolumn{1}{c}{Bottle} & 99.2/99.8/98.1/93.9 & 99.3/99.8/97.7/91.8 & 99.9/100.0/98.7/96.7 & 99.4/99.8/97.7/94.7 & 100.0/99.9/99.5/93.9 & 99.6/99.9/97.7/94.4 & 99.8/100.0/98.6/95.2 & \cellcolor{tab_ours}98.7{\footnotesize{$\pm$0.9}}/99.6{\footnotesize{$\pm$0.2}}/97.2{\footnotesize{$\pm$0.1}}/94.3{\footnotesize{$\pm$0.2}} \\

\multicolumn{1}{c}{} & \multicolumn{1}{c}{\cellcolor{tab_others}Cable} & \cellcolor{tab_others}91.0/95.1/96.4/88.5 & \cellcolor{tab_others}88.4/92.9/94.3/74.7 & \cellcolor{tab_others}94.0/98.2/96.9/93.6 & \cellcolor{tab_others}90.4/93.9/93.2/84.2 & \cellcolor{tab_others}99.9/96.9/97.6/87.8 & \cellcolor{tab_others}91.2/95.0/92.4/81.7  & \cellcolor{tab_others}92.1/95.8/95.3/89.5 & \cellcolor{tab_ours}94.7{\footnotesize{$\pm$0.7}}/97.3{\footnotesize{$\pm$0.3}}/94.0{\footnotesize{$\pm$0.5}}/96.9{\footnotesize{$\pm$0.7}} \\

\multicolumn{1}{c}{} & \multicolumn{1}{c}{Capsule} & 72.8/91.0/96.5/86.6 & 77.3/93.3/96.4/90.6 & 98.7/99.4/97.7/92.5 & 89.8/97.7/92.3/85.7 & 100.0/97.0/99.3/79.2 & 91.1/98.0/97.7/94.3 & 91.8/98.3/97.5/92.6 & \cellcolor{tab_ours}95.4{\footnotesize{$\pm$1.8}}/99.0{\footnotesize{$\pm$0.4}}/97.7{\footnotesize{$\pm$0.2}}/96.7{\footnotesize{$\pm$0.2}} \\
\bottomrule
\end{tabular}
}
\label{tab:mvtec-2-shot}%
\end{table*}%

\begin{table*}[!h]
\centering
\caption{Subset-wise performance comparison results of the 4-shot setting for AUROC/AP/pAUROC/PRO on MVTecAD.}
\setlength{\tabcolsep}{0.5mm}
\resizebox{\linewidth}{!}{
\begin{tabular}{ccccccccccc}
\toprule
\multicolumn{2}{c}{Method~$\rightarrow$} & PatchCore~\cite{roth2022towards} & WinCLIP~\cite{jeong2023winclip} & APRIL-GAN~\cite{chen2023zero} & AnomalyGPT~\cite{gu2024anomalygpt} & PromptAD~\cite{li2024promptad} & KAG-prompt~\cite{tao2024kernel} & One-for-All~\cite{lvone} & \cellcolor{tab_ours}VisionAD \\
\cline{1-2}
\multicolumn{2}{c}{Category~$\downarrow$} & CVPR'22 & CVPR'23 & arXiv'23 & AAAI'24 & CVPR'24 & AAAI'25 & ICLR'25 & \cellcolor{tab_ours}Ours \\
\hline
\multicolumn{1}{c}{\multirow{10}[1]{*}{\begin{turn}{-90}Objects\end{turn}}} & \multicolumn{1}{c}{Bottle} & 99.2/99.8/98.2/94.0 & 99.3/99.8/97.8/91.6 & 99.9/100.0/98.7/96.6 & 99.4/99.8/98.2/95.2 & 100.0/100.0/99.5/94.5 & 99.5/99.9/98.0/94.6 & 99.1/99.7/98.6/95.5 & \cellcolor{tab_ours}99.3{\footnotesize{$\pm$0.8}}/99.8{\footnotesize{$\pm$0.2}}/97.5{\footnotesize{$\pm$0.1}}/94.6{\footnotesize{$\pm$0.2}} \\

\multicolumn{1}{c}{} & \multicolumn{1}{c}{\cellcolor{tab_others}Cable} & \cellcolor{tab_others}91.0/97.1/97.5/91.7 & \cellcolor{tab_others}90.9/94.4/94.9/77.0 & \cellcolor{tab_others}94.2/98.3/97.2/94.3 & \cellcolor{tab_others}91.4/95.4/94.3/85.6 & \cellcolor{tab_others}98.8/97.4/98.2/88.9 & \cellcolor{tab_others}91.5/95.3/92.5/82.0  & \cellcolor{tab_others}95.4/97.5/96.4/91.2 & \cellcolor{tab_ours}94.8{\footnotesize{$\pm$1.5}}/97.4{\footnotesize{$\pm$0.6}}/94.2{\footnotesize{$\pm$0.4}}/87.5{\footnotesize{$\pm$0.5}} \\

\multicolumn{1}{c}{} & \multicolumn{1}{c}{Capsule} & 72.8/94.9/96.8/87.8 & 82.3/95.1/96.2/90.1 & 98.8/99.4/97.7/92.6 & 91.1/98.5/93.4/87.1 & 100.0/98.6/99.3/88.7 & 92.9/98.4/97.9/94.9 & 94.5/98.9/97.2/88.9 & \cellcolor{tab_ours}96.9{\footnotesize{$\pm$0.3}}/99.3{\footnotesize{$\pm$0.0}}/98.0{\footnotesize{$\pm$0.1}}/97.1{\footnotesize{$\pm$0.1}} \\
\bottomrule
\end{tabular}
}
\label{tab:mvtec-4-shot}%
\end{table*}%

\begin{table*}[!h]
\centering
\caption{Subset-wise performance comparison results of the 1-shot setting for AUROC/AP/pAUROC/PRO on VisA.}
\setlength{\tabcolsep}{1.2mm}
\resizebox{\linewidth}{!}{
\begin{tabular}{ccccccccccc}
\toprule
Method~$\rightarrow$ & PatchCore~\cite{roth2022towards} & WinCLIP~\cite{jeong2023winclip} & APRIL-GAN~\cite{chen2023zero} & AnomalyGPT~\cite{gu2024anomalygpt} & PromptAD~\cite{li2024promptad} & KAG-prompt~\cite{tao2024kernel} & One-for-All~\cite{lvone} & \cellcolor{tab_ours}VisionAD \\
\cline{1-1}
Category~$\downarrow$ & CVPR'22 & CVPR'23 & arXiv'23 & AAAI'24 & CVPR'24 & AAAI'25 & ICLR'25 & \cellcolor{tab_ours}Ours \\
\midrule

pcb1 & 78.9/72.1/96.1/79.0 & 75.6/76.5/96.6/82.5 & 89.0/90.2/95.3/90.4 & 82.6/76.0/97.9/83.1 & 88.7/88.0/96.6/89.2 & 84.4/81.0/97.4/85.8 & 83.5/77.4/98.5/84.3 & \cellcolor{tab_ours}93.7{\footnotesize{$\pm$0.9}}/92.7{\footnotesize{$\pm$1.2}}/98.8{\footnotesize{$\pm$0.1}}/90.1{\footnotesize{$\pm$0.6}}  \\

\cellcolor{tab_others}pcb2 & \cellcolor{tab_others}81.5/84.4/95.4/80.9 & \cellcolor{tab_others}62.2/64.9/93.0/73.6 & \cellcolor{tab_others}85.4/87.1/92.5/79.2 & \cellcolor{tab_others}75.6/76.9/92.5/64.9 & \cellcolor{tab_others}71.6/75.4/93.5/79.3 & \cellcolor{tab_others}83.3/86.8/94.0/67.7 & \cellcolor{tab_others}72.6/70.2/94.6/77.3 &\cellcolor{tab_ours}92.9{\footnotesize{$\pm$2.1}}/93.4{\footnotesize{$\pm$2.0}}/95.8{\footnotesize{$\pm$0.5}}/83.1{\footnotesize{$\pm$2.6}}  \\

pcb3 & 82.7/84.6/96.2/78.1 & 74.1/73.5/94.3/79.5 & 87.8/89.4/93.2/81.3 & 75.0/79.9/94.6/69.9 & 97.1/75.2/95.9/84.0 & 81.4/83.2/95.6/75.7 & 71.8/73.2/93.0/76.8 & \cellcolor{tab_ours}94.4{\footnotesize{$\pm$2.4}}/94.8{\footnotesize{$\pm$2.1}}/97.2{\footnotesize{$\pm$0.6}}/88.3{\footnotesize{$\pm$1.9}}  \\
\bottomrule
\end{tabular}
}
\label{tab:visa-1-shot}%
\end{table*}%

\begin{table*}[!h]
\centering
\caption{Subset-wise performance comparison results of the 2-shot setting for AUROC/AP/pAUROC/PRO on VisA.}
\setlength{\tabcolsep}{1.2mm}
\resizebox{\linewidth}{!}{
\begin{tabular}{ccccccccccc}
\toprule
Method~$\rightarrow$ & PatchCore~\cite{roth2022towards} & WinCLIP~\cite{jeong2023winclip} & APRIL-GAN~\cite{chen2023zero} & AnomalyGPT~\cite{gu2024anomalygpt} & PromptAD~\cite{li2024promptad} & KAG-prompt~\cite{tao2024kernel} & One-for-All~\cite{lvone} & \cellcolor{tab_ours}VisionAD \\
\cline{1-1}
Category~$\downarrow$ & CVPR'22 & CVPR'23 & arXiv'23 & AAAI'24 & CVPR'24 & AAAI'25 & ICLR'25 & \cellcolor{tab_ours}Ours \\
\midrule

pcb1 & 71.5/72.6/97.6/86.1 & 76.7/78.3/97.0/83.8 & 90.9/91.6/95.8/90.6 & 83.6/78.7/96.5/82.9 & 90.9/80.9/96.9/90.2 & 88.8/86.7/98.0/87.0 & 86.5/78.5/98.5/85.1 & \cellcolor{tab_ours}93.3{\footnotesize{$\pm$0.6}}/92.1{\footnotesize{$\pm$0.9}}/98.8{\footnotesize{$\pm$0.1}}/90.1{\footnotesize{$\pm$0.5}}  \\

\cellcolor{tab_others}pcb2 & \cellcolor{tab_others}84.3/86.6/96.0/82.9 & \cellcolor{tab_others}62.6/65.8/93.9/76.2 & \cellcolor{tab_others}89.2/91.3/92.8/79.7 & \cellcolor{tab_others}78.3/75.8/93.0/64.8 & \cellcolor{tab_others}73.0/73.0/94.8/79.3 & \cellcolor{tab_others}82.0/85.2/95.2/73.9 & \cellcolor{tab_others}75.2/73.6/96.8/77.8 & \cellcolor{tab_ours}93.3{\footnotesize{$\pm$0.4}}/93.4{\footnotesize{$\pm$0.9}}/96.0{\footnotesize{$\pm$0.1}}/83.6{\footnotesize{$\pm$0.8}}  \\

pcb3 & 84.8/86.1/97.1/82.2 & 78.8/80.9/95.1/82.3 & 90.9/92.2/93.5/81.5 & 79.2/80.4/95.9/72.4 & 76.2/82.8/96.1/84.7 & 89.0/90.0/95.6/75.5 & 70.7/74.3/93.7/78.7 & \cellcolor{tab_ours}93.5{\footnotesize{$\pm$3.5}}/94.2{\footnotesize{$\pm$2.6}}/97.4{\footnotesize{$\pm$0.3}}/89.2{\footnotesize{$\pm$1.5}} \\
\bottomrule
\end{tabular}
}
\label{tab:visa-2-shot}%
\end{table*}%

\begin{table*}[!h]
\centering
\caption{Subset-wise performance comparison results of the 4-shot setting for AUROC/AP/pAUROC/PRO on VisA.}
\setlength{\tabcolsep}{1.2mm}
\resizebox{\linewidth}{!}{
\begin{tabular}{ccccccccccc}
\toprule
Method~$\rightarrow$ & PatchCore~\cite{roth2022towards} & WinCLIP~\cite{jeong2023winclip} & APRIL-GAN~\cite{chen2023zero} & AnomalyGPT~\cite{gu2024anomalygpt} & PromptAD~\cite{li2024promptad} & KAG-prompt~\cite{tao2024kernel} & One-for-All~\cite{lvone} & \cellcolor{tab_ours}VisionAD \\
\cline{1-1}
Category~$\downarrow$ & CVPR'22 & CVPR'23 & arXiv'23 & AAAI'24 & CVPR'24 & AAAI'25 & ICLR'25 & \cellcolor{tab_ours}Ours \\
\midrule

pcb1 & 84.7/81.0/98.1/88.1 & 88.3/87.7/98.1/87.9 & 91.2/91.8/96.0/90.6 & 83.1/77.1/96.8/83.1 & 90.9/90.1/98.2/90.2 & 89.7/87.6/97.9/87.2 & 85.2/78.6/98.7/85.6 & \cellcolor{tab_ours}93.8{\footnotesize{$\pm$0.7}}/92.9{\footnotesize{$\pm$1.4}}/98.9{\footnotesize{$\pm$0.1}}/90.9{\footnotesize{$\pm$0.3}}  \\

\cellcolor{tab_others}pcb2 & \cellcolor{tab_others}84.3/86.2/96.6/83.7 & \cellcolor{tab_others}67.5/71.3/94.6/78.0 & \cellcolor{tab_others}89.2/91.3/93.0/80.2 & \cellcolor{tab_others}85.5/86.7/93.2/71.3 & \cellcolor{tab_others}78.6/75.3/95.3/76.3 & \cellcolor{tab_others}87.8/90.7/96.1/77.7 & \cellcolor{tab_others}75.5/73.9/96.6/77.4 & \cellcolor{tab_ours}94.4{\footnotesize{$\pm$0.9}}/94.7{\footnotesize{$\pm$1.0}}/96.3{\footnotesize{$\pm$0.1}}/85.0{\footnotesize{$\pm$0.9}}  \\

pcb3 & 87.0/88.3/97.4/84.4 & 83.3/84.8/95.8/84.2 & 91.6/92.7/93.7/81.5 & 86.9/88.3/96.1/73.1 & 80.3/83.5/96.8/85.0 & 85.8/86.9/96.4/78.6 & 74.7/74.6/94.0/78.1 & \cellcolor{tab_ours}96.7{\footnotesize{$\pm$0.6}}/96.9{\footnotesize{$\pm$0.5}}/97.9{\footnotesize{$\pm$0.1}}/90.3{\footnotesize{$\pm$0.7}}  \\
\bottomrule
\end{tabular}
}
\label{tab:visa-4-shot}%
\end{table*}%

\begin{table*}[t]
  \centering
    \caption{Subset-wise performance comparison on VisA under 1-/2-/4-shot settings with AUROC, AP, pAUROC, and PRO metrics.}
    \setlength{\tabcolsep}{1.0mm}
    \resizebox{0.95\linewidth}{!}{
    \begin{tabular}{ccccccccccccc}
    \toprule
    \multirow{2}{*}{Category} & \multicolumn{4}{c}{1-shot} & \multicolumn{4}{c}{2-shot} & \multicolumn{4}{c}{4-shot} \\
\cmidrule(r){2-5} \cmidrule(r){6-9} \cmidrule(r){10-13}  & AUROC & AUPR & pAUROC & PRO & AUROC & AUPR & pAUROC & PRO & AUROC & AUPR & pAUROC & PRO \\

    \midrule
    audiojack & 56.3{\footnotesize{$\pm$3.1}}   & 41.3{\footnotesize{$\pm$5.2}} & 94.7{\footnotesize{$\pm$0.5}} & 77.5{\footnotesize{$\pm$1.4}} & 59.8{\footnotesize{$\pm$2.3}} & 40.8{\footnotesize{$\pm$1.2}} & 95.7{\footnotesize{$\pm$0.8}} & 81.0{\footnotesize{$\pm$3.1}} & 68.6{\footnotesize{$\pm$5.3}} & 51.0{\footnotesize{$\pm$9.7}} & 97.2{\footnotesize{$\pm$0.7}} & 86.7{\footnotesize{$\pm$2.9}}  \\
    
    \cellcolor{tab_others}bottle cap & \cellcolor{tab_others}75.8{\footnotesize{$\pm$0.8}}  & \cellcolor{tab_others}62.6\footnotesize{$\pm$1.0} & \cellcolor{tab_others}98.4\footnotesize{$\pm$0.2} & \cellcolor{tab_others}92.2\footnotesize{$\pm$0.8} & \cellcolor{tab_others}79.4\footnotesize{$\pm$4.5} & \cellcolor{tab_others}68.8\footnotesize{$\pm$8.6} & \cellcolor{tab_others}98.7\footnotesize{$\pm$0.5} & \cellcolor{tab_others}93.4\footnotesize{$\pm$2.4} & \cellcolor{tab_others}83.0\footnotesize{$\pm$4.4} & \cellcolor{tab_others}76.0\footnotesize{$\pm$9.1} & \cellcolor{tab_others}99.1\footnotesize{$\pm$0.6} & \cellcolor{tab_others}95.4\footnotesize{$\pm$3.0}  \\
    
    button battery & 77.8\footnotesize{$\pm$2.4}    & 83.7\footnotesize{$\pm$1.6}   & 96.2\footnotesize{$\pm$0.4}   & 85.8\footnotesize{$\pm$0.9}   & 79.2\footnotesize{$\pm$1.8}   & 84.5\footnotesize{$\pm$1.0}   & 96.6\footnotesize{$\pm$0.2}   & 86.6\footnotesize{$\pm$1.6}   & 81.2\footnotesize{$\pm$2.9}   & 86.0\footnotesize{$\pm$2.3}   & 97.2\footnotesize{$\pm$0.5}   & 87.8\footnotesize{$\pm$3.0}  \\
\bottomrule
    \end{tabular}%
    }
  \label{tab:real-iad}
\end{table*}

\end{document}